\documentclass[runningheads]{llncs}

\usepackage[T1]{fontenc}
\usepackage{hyperref}
\usepackage[misc]{ifsym}

\usepackage{amssymb}
\usepackage{amsmath}
\usepackage{algorithm}
\usepackage[noend]{algpseudocode}
\usepackage{nth}
\usepackage[all]{nowidow}
\usepackage{booktabs,etoolbox}
\usepackage{multirow}
\usepackage{caption}
\usepackage{graphicx}
\usepackage{subcaption}
\usepackage{overpic}
\usepackage{makecell}
\usepackage{enumitem}

\usepackage{tikz}
\usetikzlibrary{fit}


\usepackage{array}
\usepackage{ragged2e}
\newcolumntype{P}[1]{>{\centering\arraybackslash}p{#1}}
\newcolumntype{R}[1]{>{\RaggedLeft\arraybackslash}p{#1}}

\usepackage[export]{adjustbox}
\usepackage{arydshln}

\newcommand{\overbar}[1]{\mkern 1.mu\overline{\mkern-1.mu#1\mkern-1.mu}\mkern 1.mu}

\usepackage{pifont}

\usepackage{amsmath}

\newcommand{\Method}{\text{VideoPCDNet}}
\newcommand{\SpaceInvaders}{\text{Space-Invaders}}

\newcommand{\NumContext}{C}

\newcommand{\ImageRange}[2]{\mathbf{X}_{#1:{#2}}}

\newcommand{\Image}{\textbf{X}}

\newcommand{\ImageT}[1]{\textbf{X}_{#1}}

\newcommand{\Objects}{{\mathcal{O}}}

\newcommand{\Object}[1]{\textbf{O}_{#1}}

\newcommand{\NumObjects}{N}

\newcommand{\PredObject}[1]{\hat{\textbf{O}}_{#1}}

\newcommand{\Prototypes}{\mathcal{P}}
\newcommand{\Prototype}[1]{\textbf{P}_{#1}}
\newcommand{\Masks}{\mathcal{M}}
\newcommand{\Mask}[1]{\textbf{M}_{#1}}
\newcommand{\NumPrototypes}{P}

\newcommand{\LocalizationMatrix}{\textbf{L}}
\newcommand{\FFT}{\mathcal{F}}
\newcommand{\IFFT}{\mathcal{F}^{-1}}
\newcommand{\Conj}[1]{\overbar{#1}}
\newcommand{\PhaseDif}{\Delta \theta}

\newcommand{\Templates}{\mathcal{T}}
\newcommand{\Template}[1]{\textbf{T}_{#1}}

\newcommand{\Compose}{\mathcal{G}}

\newcommand{\SceneState}{\textbf{s}}

\newcommand{\Color}{\text{\textbf{c}}}
\newcommand{\CoM}{\text{\textbf{Z}}}
\newcommand{\ProtoID}{\text{\textbf{P}}}

\newcommand{\Velocity}{\textbf{v}}

\newcommand{\Figure}[1]{Fig.~\ref{#1}}

\newcommand{\Figuress}[2]{Figs.~\ref{#1}--\ref{#2}}

\newcommand{\Equation}[1]{Eq.~\eqref{#1}}

\newcommand{\Section}[1]{Sec.~\ref{#1}}

\newcommand{\Appendix}[1]{Appendix~\ref{#1}}

\newcommand{\Algorithm}[1]{Algorithm~\ref{#1}}

\begin{document}
\title{\Method: Video Parsing and Prediction with Phase Correlation Networks}
\titlerunning{Video Parsing and Prediction with Phase Correlation Networks}

\author{Noel José Rodrigues Vicente\inst{1}, Enrique Lehner\inst{1}, Angel Villar-Corrales\inst{1,2,3}\Letter,\\ Jan Nogga\inst{1,2,3} \and Sven Behnke\inst{1,2,3}}
\authorrunning{Rodrigues et al.}
\institute{
	Autonomous Intelligent Systems group, University of Bonn, Germany
	\and 
	Lamarr Institute for Machine Learning and Artificial Intelligence
	\and
	Center for Robotics, University of Bonn, Germany
	\\
	\email{villar@ais.uni-bonn.de}
}

\maketitle              

\begin{abstract}
\vspace{-0.23cm}
Understanding and predicting video content is essential for planning and reasoning in dynamic environments.
Despite advancements, unsupervised learning of object representations and dynamics remains challenging.
We present \Method, an unsupervised framework for object-centric video decomposition and prediction.
Our model uses fre\-quency-domain phase correlation techniques to recursively parse videos into object components, which are represented as transformed versions of learned object prototypes, enabling accurate and interpretable tracking.
By explicitly modeling object motion through a combination of frequency domain operations and lightweight learned modules, \Method{} enables accurate unsupervised object tracking and prediction of future video frames.
In our experiments, we demonstrate that \Method{} outperforms multiple object-centric baseline models for unsupervised tracking and prediction on several synthetic datasets, while learning interpretable object and motion representations.

\vspace{-0.15cm}
\end{abstract}

\section{Introduction}
Humans naturally interpret dynamic scenes by segmenting them into discrete objects and tracking their interactions over time.
Recent works in object-centric video prediction imitate this cognitive process by learning unsupervised object representations and modeling their dynamics and interactions using recurrent neural networks~\cite{Zoran_PARTS_2021} (RNNs) or transformers~\cite{Villar_OCVP_2023,Wu_SlotFormer_2023,Daniel_DDLP_2024}.
However, these methods typically incur high computational costs, require large amounts of training data, and produce models whose internal representations remain largely opaque.

In this work, we propose \Method, an unsupervised video decomposition and prediction model that extends the Phase-Correlation Decomposition Network (PCDNet)~\cite{Villar_PCDNet_2022} to the video domain.
Unlike conventional object-centric methods, which rely on high-dimensional latent spaces to capture object representations and dynamics, our proposed \Method{} represents objects as transformed versions of a set of learned object prototypes and explicitly encodes their motion using phase differences.
This design not only enables efficient object tracking and scene parsing, but also yields representations that are inherently interpretable.
Our model recursively parses a video sequence into its object components, leading to an accurate and robust object tracking even under challenging motion dynamics and occlusions.
Furthermore, \Method{} explicitly models object motion by combining phase-correlation techniques with lightweight learned modules in order to efficiently forecast future object states and video frames with a minimal number of trainable parameters.

In our experiments, we demonstrate that \Method{} outperforms multiple baseline models for unsupervised object tracking and future frame video prediction on several synthetic datasets while also learning interpretable object and motion representations.
Our work thus demonstrates that integrating frequency-domain processing with object-centric representation learning can yield a more efficient and interpretable framework for video prediction.

\section{Related Work}

\subsection{Object-Centric Learning}

\paragraph{Object-centric representation learning.} Object-centric learning methods aim to decompose an image or video into a set of object components in an unsupervised manner.
These objects can be represented as unconstrained latent vectors (often called slots)~\cite{Burgess_MonetUnsupervisedSceneDecompositionRepresentation_2019,Weis_UnmaskingInductiveBiasesOfUnsupervisedObjectRepresentationsForVideoSequences_2020,Locatello_ObjectCentricLearningWithSlotAttention_2020,Jiang_ScalorGenerativeWorldModelsWithScalableObjectRepresentations_2019,Kipf_ConditionalObjectCentricLearningFromVideo_2022}, factored latent variables~\cite{He_TrackingByAnnimation_2019,Stanic_HierarchicalRelationalInference_2020}, spatial mixture models~\cite{Greff_IodineMultiObjectRepresentationLearningWithIterativeVariationalInference_2019,Engelcke_GenesisV2InferringObjectRepresentationsWithoutIterativeRefinement_2021}, or explicit object prototypes~\cite{Villar_PCDNet_2022,Monnier_UnsupervisedLayeredImageDecompositionIntoObjectPrototypes_2021}.
The learned object representations benefit multiple downstream tasks, such as learning behaviors for robotic manipulation~\cite{Mosbach_SOLDReinforcementLearningSlotObjectCentricLatentDynamics,Villar_PlaySlot_2025} and unsupervised segmentation~\cite{Kipf_ConditionalObjectCentricLearningFromVideo_2022,Engelcke_GenesisV2InferringObjectRepresentationsWithoutIterativeRefinement_2021}.
\vspace{-0.3cm}

\paragraph{Object-centric video prediction.} Object-centric video prediction aims to model object dynamics and interactions to forecast future object states and frames.
Recently, several methods propose to model object dynamics using slot-based representations and different architectural priors, including RNNs~\cite{Zoran_PARTS_2021,Nakano_InteractionBasedDisentanglement_2023} or transformers~\cite{Villar_OCVP_2023,Wu_SlotFormer_2023,Daniel_DDLP_2024}.
In contrast, \Method{} leverages phase-correlation as a strong inductive bias to model object motion interpretably and with minimal learnable parameters.

\subsection{Phase-Correlation Networks}

Phase-correlation networks are a class of neural networks that incorporate the differentiable phase-correlation technique~\cite{Alba_PahseCorrelationImageAlignment_2012} to estimate transformation parameters, such as translations, between two signals by analyzing the phase differences in their Fourier transforms.
Integrating this operation into neural networks often leads to interpretable and compact models for image and video tasks.
PCDNet~\cite{Villar_PCDNet_2022} uses the phase correlation method to align a set of learned object prototypes with input images, enabling unsupervised object-centric image decomposition.
In the video domain, Frequency Domain Transformer Networks (FDTN)~\cite{Farazi_FDTN_2019} compute phase differences between consecutive video frames to model linear motion, and recursively apply the inferred motion to forecast future video frames.
Several works extended this approach for motion segmentation~\cite{Farazi_MotionSegmentationUsingFrequencyDomainTransformerNetworks_2020,Farazi_LocalFrequencyDomainTransformerNwtworksForVideoPrediction_2021}, modeling object rotations and scale variations~\cite{Wolter_ObjectCenteredFourierMotionEstimation_2020}, stochastic video prediction~\cite{Farazi_IntentionAwareFrequencyDomainNetworks_2022} or learning relational motion~\cite{Mosbach_Fourier_2021}.
Unlike previous methods, which treat the scene's motion holistically or merely separate foreground from background, our proposed \Method{} parses a video sequence into individual object components and explicitly models each object's motion.

\section{\Method}

\begin{figure}[t]
	\centering 
	\begin{subfigure}[b]{0.49\linewidth}
		\label{fig:fig} \footnotesize
		\includegraphics[width=\linewidth]{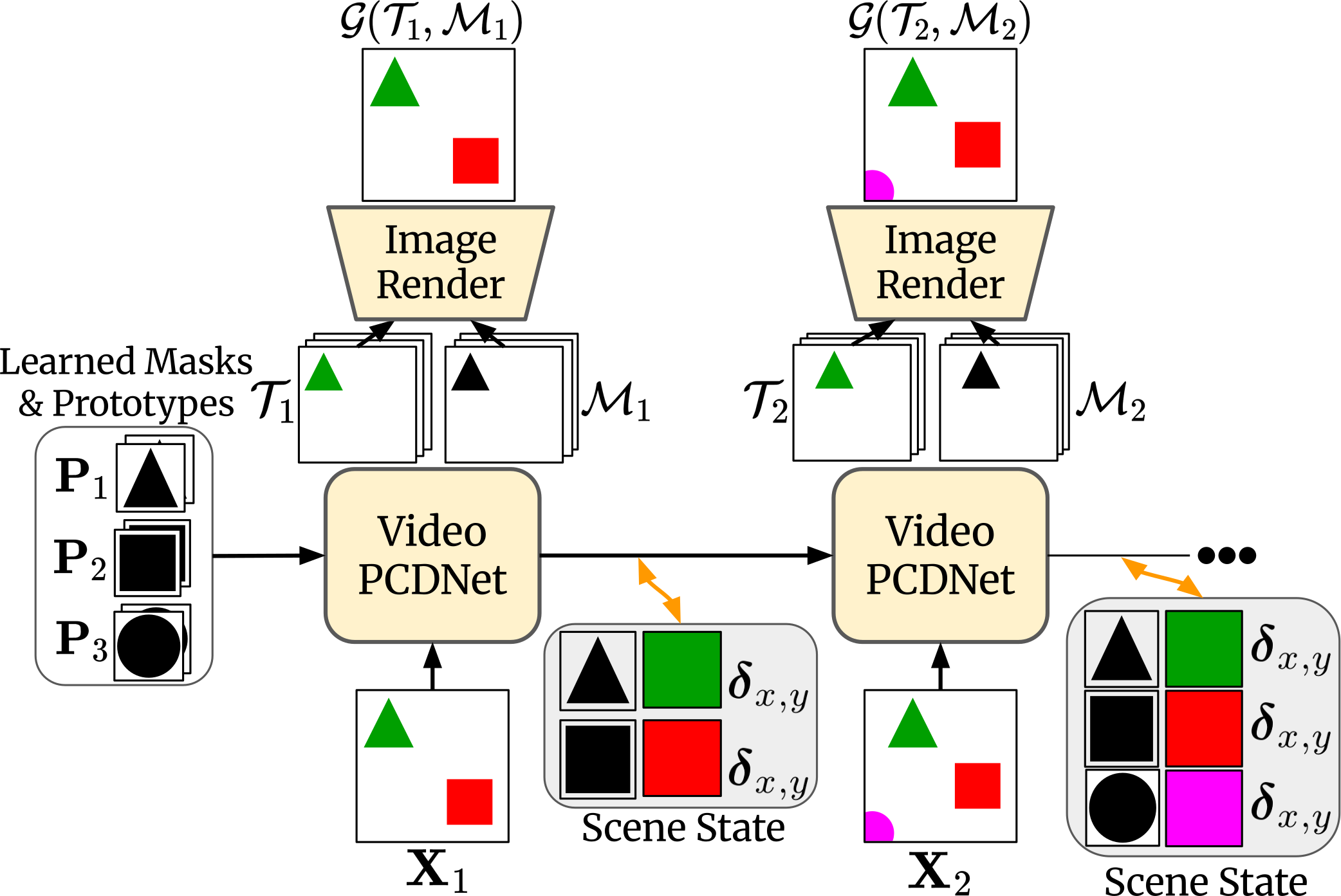}
		\caption{\Method{} Pipeline}
	\end{subfigure}
	\hfill
	\begin{subfigure}[b]{0.49\linewidth}
		\label{fig:fig2} \footnotesize
		~~\includegraphics[width=\linewidth]{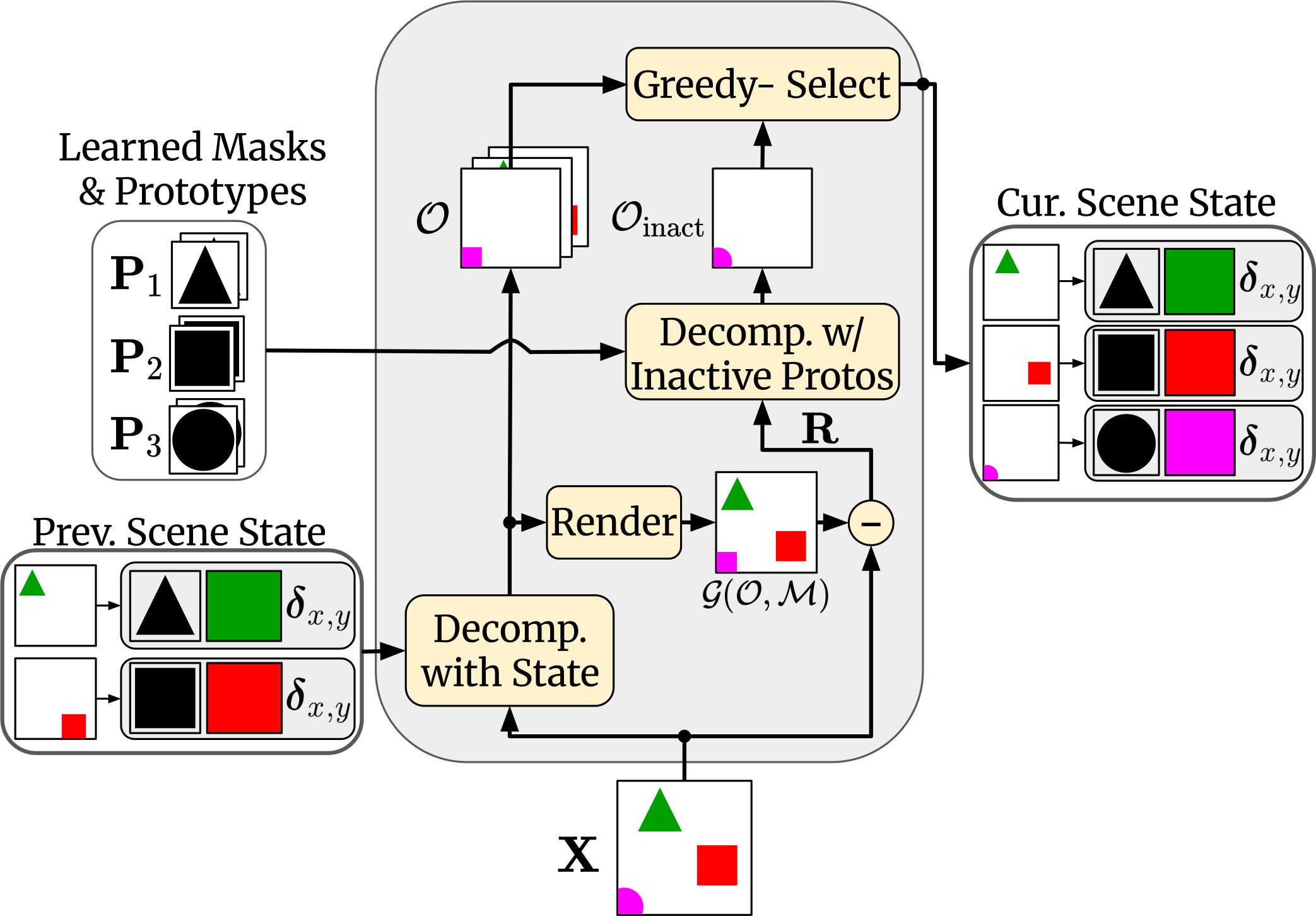}
		\caption{\Method{} Object Parsing}
	\end{subfigure}
	\vspace{-0.25cm}
	\caption{
		Overview of \Method.
		Our model recursively parses a video sequence $\ImageT{i}$ into its object components, which are represented as transformed versions of a set of learned object prototypes $\Prototypes$.
		\Method{} maintains an interpretable scene state, which encodes the objects present in the scene and their properties, thus enabling interpretable object prediction and tracking.
	}
	\label{fig:main}
	\vspace{-0.5cm}
\end{figure}

We propose \Method, illustrated in~\Figure{fig:main}, a novel framework that extends the phase-correlation decomposition network PCDNet~\cite{Villar_PCDNet_2022} for object-centric video parsing and prediction.
Given a sequence of images $\ImageRange {1}{\NumContext}$, our method recursively parses each frame into $\NumObjects$ independent object components $\Objects = \{\Object{1}, ..., \Object{\NumObjects}\}$, where each of these objects corresponds to a transformed version of a prototype from a set of learned object prototypes $\Prototypes = \{\Prototype{1}, ..., \Prototype{\NumPrototypes}\}$.
To allow for a temporally consistent object-centric representation, each object $\Object{i}$ is represented by a state $\SceneState{}$ that encodes interpretable attributes such as shape, color, and position, enabling robust object tracking over time (\Section{sec: VideoProcessing}).
A key component in \Method{} is the \emph{Motion Module} (\Section{sec: MotionModule}), which leverages phase correlation in the frequency domain to predict future object states, which can be used to render a video frames.

\subsection{Preliminaries: PCDNet}
\label{sec: PCDNet}

PCDNet~\cite{Villar_PCDNet_2022} is an unsupervised decomposition method that parses images into distinct object-centric components, which are represented as transformed versions of a set of learned object prototypes $\Prototypes$.
Each object prototype $\Prototype{j}$ is learned along with its corresponding mask $\Mask{j}$, which is used to model depth ordering and occlusions.
At its core, PCDNet leverages a differentiable phase-correlation mechanism, known as PC-Cell, to align learned prototypes with objects in the input image $\Image$ by estimating spatial translations in the frequency domain.
Specifically, the PC-Cell computes the cross-correlation in the frequency domain between an object prototype and the input image, producing a localization matrix $\LocalizationMatrix$ that encodes potential object locations as correlation peaks:
\vspace{-0.5cm}
\begin{align}
\begin{minipage}{0.45\linewidth}
	\begin{equation}
		\PhaseDif = \left( \frac{\FFT(\Image) \odot \Conj{\FFT(\Prototype{})}}
		{|| \FFT(\Image) \odot \Conj{\FFT(\Prototype{})} ||} \right),
		\label{eq: pc}
	\end{equation}
\end{minipage}
\hfill
\begin{minipage}{0.35\linewidth}
	\begin{equation}
		\LocalizationMatrix = \IFFT(\PhaseDif),
		\label{eq: loc}
	\end{equation}
\end{minipage}
\end{align}
where $\FFT$ and $\IFFT$ denote the Fourier and Inverse Fourier transforms, respectively, and $\Conj{\FFT(\Prototype{})}$ denotes the complex conjugate of $\FFT(\Prototype{})$.
From this matrix, the most prominent peaks are extracted, which represent the spatial shifts $(\delta_x, \delta_y)$ that best align the corresponding prototype $\Prototype{}$ to the scene.

Originally, PCDNet handles color information via a separate Color Module, which adjusts prototype colors only after spatial alignment, disregarding color cues during the decomposition process.
In contrast, our framework incorporates color as an additional feature for decomposition.
To achieve this, we discretize the image colors using $k$-means clustering, converting the image into a multi-channel representation where each channel corresponds to distinct color cluster.

Our refined PC-Cell operates by computing channel-wise phase-correlation (\Equation{eq: pc}) between the object prototypes and this multi-channel image representation.
%
Each correlation peak is thus represented by a triplet \((\Color, \delta_x, \delta_y)\), identifying the strongest responding color channel \Color{} and corresponding shift parameters \((\delta_x, \delta_y)\).
Using the Fourier shift theorem, the transformed prototypes $\Templates=\{\Template{1}, ... \Template{|\Templates|}\}$, denoted as object templates, are computed as:
\begin{align}
& \Template{} = \IFFT \left( \FFT(\Prototype{}) \cdot \exp (-i2\pi(\delta_x \mathbf{f}_x + \delta_y \mathbf{f}_y)) \right),
\end{align}
where $\mathbf{f}_x$ and $\mathbf{f}_y$  denote the frequencies along the
horizontal and vertical directions, respectively.

Finally, PCDNet employs an iterative greedy algorithm, described in \Algorithm{alg:greedy}, in order to select the transformed prototypes $\Templates$ and their corresponding transformed masks $\Masks$ that best represent the image.
This greedy algorithm iteratively selects the object template that, combined with the previously selected templates, minimizes the reconstruction error using \Equation{eq: err}.
The object templates and masks are composed using \Equation{eq: compose} to reconstruct the images, such that the first selected object ($\Template{1}$) corresponds to the one closest to the viewer, whereas the last selected object ($\Template{\NumObjects}$) is located the furthest from the viewer; thus inherently modeling relative depth ordering between objects.
\begin{align}
	& \textbf{E}(\Image, \Templates) = || \Image - \Compose(\Templates, \Masks) ||,	\label{eq: err}
	\\
	&\Compose(\Templates,\Masks) = \Template{i+1} \odot (1 - \Mask{i}) + \Template{i} \odot \Mask{i} \;\; \forall \; i \in \{\NumObjects, ..., 1\}.
	\label{eq: compose}
\end{align}

PCDNet is trained end-to-end by minimizing an image reconstruction error, as well as two regularization costs to enforce sparsity in the object prototypes and smooth object masks.

\begin{algorithm}[t]
	\caption{Greedy Object and Mask Selection for Reconstruction}
	\label{alg:greedy}
	\begin{algorithmic}[1]
		\Procedure{GreedySelect}{$\mathbf{X}, \mathcal{T}, \mathcal{M}, \texttt{max\_objs}$}
		\State \(\mathcal{S} \gets \varnothing\)
		\While{\(|\mathcal{S}| < \texttt{max\_objs}\)}
		\ForAll{\(j \notin \mathcal{S}\)}
		\State Let \(\mathcal{S}' \gets \mathcal{S} \cup \{j\}\)
		\State Compute the reconstruction $\Compose(\Templates_{\mathcal{S}'},\ \Masks_{\mathcal{S}'})$ (\Equation{eq: compose})
		\State Compute error $\textbf{E}_j = || \textbf{X} - \Compose(\Templates_{\mathcal{S}'},\ \Masks_{\mathcal{S}'})||$  (\Equation{eq: err})
		\EndFor
		\State \(j^* \gets \arg\min_{j \notin \mathcal{S}} \textbf{E}_j\)
		\State \(\mathcal{S} \gets \mathcal{S} \cup \{j^*\}\)
		\EndWhile
		\State \Return \(\mathcal{T}_{\mathcal{S}},\; \mathcal{M}_{\mathcal{S}}\)
		\EndProcedure
	\end{algorithmic}
\end{algorithm}
\vspace{-0.5cm}

\subsection{Video Processing with PCDNet}
\label{sec: VideoProcessing}

\subsubsection{State Representation and Alignment}
\label{sec: state}

While PCDNet successfully parses images into their object components, extending this algorithm for video processing introduces several challenges, such as consistent object tracking.

Our proposed \Method{} addresses these issues by recursively tracking and updating a consistent scene state that determines the objects present in the scene, as well as their main attributes.
Namely, \Method{} computes for each object $k$ in the scene a state $\SceneState_k = (\Color_k, \ProtoID_k,\CoM_k)$ that represents the object color $\Color_k$, appearance $\ProtoID_k$, and center of mass $\CoM_k$.
This state representation enables \Method{} to obtain a temporally consistent tracking of the objects in the scene by conditioning and aligning the decomposition process with the current state.
Given the object representations from a video frame and the current \Method{} state, we define the cost of matching each parsed object $i$ to each tracked object $j$ as:
\begin{align}
	C_{i,j} = \lambda_{\Color}\ ||\Color_i - \Color_j\|| + \lambda_{\ProtoID} ||\ProtoID_i - \ProtoID_j|| + \lambda_{\CoM} ||\CoM_i - \CoM_j||,
\end{align}
where $\lambda_{\Color}$, $\lambda_{\ProtoID}$ and $\lambda_{\CoM}$ denote weights for each representation.
The parsed objects are then aligned to the \Method{} state according to the Hungarian algorithm.
Stable identifiers are maintained by retaining the track IDs of matched objects, dropping those of unmatched ones, and assigning new IDs to newly detected objects; thus achieving temporally consistent object tracking.

\subsubsection{Two-Stage Decomposition}
\label{sec:dual-stage}

To improve the efficiency and robustness for object-centric video decomposition, \Method{} leverages a two-stage approach, which is detailed in \Algorithm{alg:dual_pass}.
In the first stage, our framework parses the current observation using the object prototypes present in the \Method{} state, reducing interference from inactive prototypes (i.e. those not present in the scene state) and improving the decomposition temporal consistency and efficiency.
However, this first stage lacks the ability to parse objects that were not represented in the state, such as objects entering the scene or previously occluded.

This limitation is addressed in the second stage, which refines the initial scene parsing if its reconstruction error exceeds a certain threshold.
The second stage performs a more exhaustive processing of the scene, where the residual error from the first stage is measured, and candidate object templates are computed using the previously inactive prototypes in order to minimize such residual.
Finally, the best object candidates among the first and second stages are selected to reconstruct the image, and aligned to update the scene state.

To further enhance robustness in challenging scenarios, such as partial occlusions, \Method{} integrates externally generated object templates ($\Templates_{ext}$) and masks ($\Masks_{ext}$) derived from predicted object states.
By concatenating these external object candidates with internal ones, \Method{} improves object decomposition consistency and accuracy, particularly in scenarios where the internal localization mechanisms alone may fail.
%
%

\begin{algorithm}[t]
\caption{Two-Stage Object-Centric Video Parsing Algorithm}
\label{alg:dual_pass}
\begin{algorithmic}[1]
\Procedure{TwoStageSelect}{$\Image,\ \Templates_{ext},\ \Masks_{ext},\ \texttt{max\_objs},\  \texttt{err\_thr},\ \Objects_{\text{prev}},\ \Prototypes}$
	\Statex \# Stage 1: Parse image with state
    \State $(\Templates, \Masks) \gets \Call{CreateCandidatesWithState}{\Image,\ \Objects_{\text{prev}}}$ 
    %
	\State $\Templates \gets \Templates \cup \Templates_{ext}, \quad \Masks \gets \Masks \cup \Masks_{ext}$
	\State $(\Objects, \Masks) \gets \Call{GreedySelect}{\Image,\ \Templates,\ \Masks,\ \texttt{max\_objs}}$
	\Statex \# Compute residual error
    \State $\mathbf{R} \gets \Image - \Compose(\Objects, \Masks)$ 
    %
    \If{\(\|\mathbf{R}\| \leq \texttt{err\_thr}\)}
	    \State \((\Objects, \Masks) \gets \Call{AlignWithPrevious}{\Objects, \Masks, \Objects_{\text{prev}}}\)
        \State \Return \((\Objects, \Masks)\)
   \EndIf
    \Statex \# Stage 2: Parse residual with inactive prototypes
 	\State $\overline{\Objects}_{\text{prev}} = \Prototypes  \setminus \Objects_{\text{prev}}$   
	\State \((\Templates_{\text{inact}}, \Masks_{\text{inact}}) \gets \Call{CreateCandidatesWithState}{\mathbf{R}, \overline{\Objects}_{\text{prev}}}\)
    \State \((\Objects_{\text{inact}}, \Masks_{\text{inact}}) \gets \Call{GreedySelect}{\mathbf{R},\ \Templates_{\text{inact}},\ \Masks_{\text{inact}},\ \texttt{max\_objs}}\)
    \State \(\Objects_{\text{all}} \gets \Objects \cup \Objects_{\text{inact}}\), \(\Masks_{\text{all}} \gets \Masks \cup \Masks_{\text{inact}}\)
    \Statex \# Select best object and mask candidates from both stages
	\State \((\Objects, \Masks) \gets \Call{GreedySelect}{\Image,\ \Objects_{\text{all}},\ \Masks_{\text{all}},\ \texttt{max\_objs}}\)
	 \Statex \# Align selected objects with state
	%
    %
    \State \((\Objects, \Masks) \gets \Call{AlignWithPrevious}{\Objects, \Masks, \Objects_{\text{prev}}}\)
    \State \Return \(\Objects, \Masks\)
\EndProcedure
\end{algorithmic}
\end{algorithm}
\vspace{-0.0cm}

\subsection{Object-Centric Motion Prediction}
\label{sec: MotionModule}

\begin{figure}[t]
	\centering
	\includegraphics[width=\textwidth]{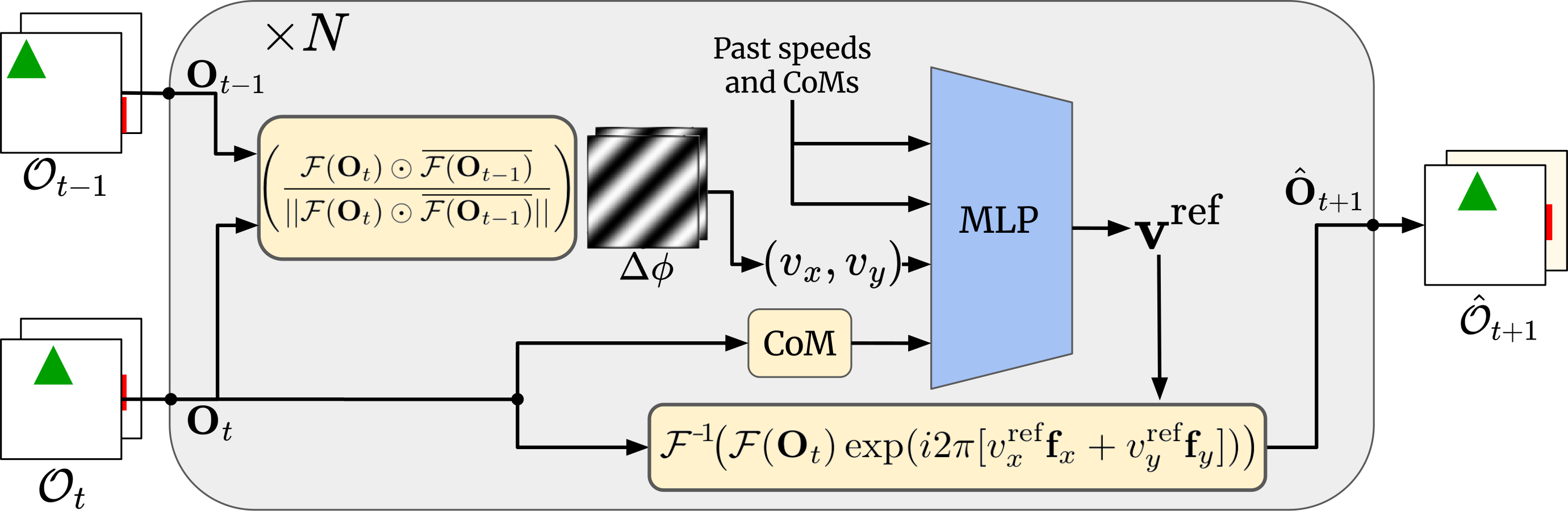}
	\vspace{-0.5cm}
	\caption{Illustration of the \emph{Motion Module} in \Method. Future object states are predicted leveraging frequency domain operations and a learned module.}
	\label{fig: motion module}
	\vspace{-0.3cm}
\end{figure}

The object-centric state used in \Method{} enables interpretable forecasting of future object states by modeling object phase differences.

Given two seed frames $\ImageT{t-1}$ and $\ImageT{t}$, we decompose these images into their aligned object components $\Objects_{t-1}$ and $\Objects_{t}$.
For each object, we compute with \Equation{eq: pc} the phase difference $\PhaseDif$ between the two time-steps, which encodes the object's translation speed.
Future object states can then be predicted by adding the estimated phase differences:
\begin{align}
	& \PredObject{t+1} = \IFFT \left( \FFT(\Object{t}) \cdot \exp(i2\pi\PhaseDif) \right).
	\label{eq: phase add}
\end{align}
Future video frames are then rendered by combing all predicted object states as described in \Section{sec: PCDNet}.

This simple, yet effective, approach for object-centric video prediction leads to temporal consistent and interpretable predictions.
However, it suffers from several limitations, such as only modeling linear motion or lacking robustness to imperfect phase differences.

To address these shortcomings, we propose to enhance our prediction approach with a learnable module, which refines the estimated object velocities, thus allowing \Method{} to accurately and robustly forecast future object states, as well as dealing with heavy occlusions and non-linear motion.

The refinement of velocities is performed by the \emph{Motion Module}, depicted in \Figure{fig: motion module}.
This module is an MLP shared among all objects in the scene, which jointly processes a temporal history of object features to produce a refined object velocity.
More precisely, the Motion Module receives the estimated object velocities and center of mass from the last time-steps, and outputs a refined velocity prediction $\Velocity^{\text{ref}} = (v^{\text{ref}}_{x}, v^{\text{ref}}_{y})$.
These refined velocities are then converted into phase differences $\PhaseDif^{\text{ref}}$:
\begin{align}
	& \PhaseDif^{\text{ref}} = 2 \pi ( v^{\text{ref}}_{x} \mathbf{f}_x + v^{\text{ref}}_{y} \mathbf{f}_y ) .
\end{align}
The future object states are then predicted through \Equation{eq: phase add} using the refined phase differences, ensuring that the model captures non-linear motion and performs a robust prediction for every object.

\section{Experiments}

\subsection{Experimental Setup}\label{sec:experimental_setup}

\noindent \textbf{Datasets: }  We evaluate \Method{} on two synthetic datasets with varied object appearance and motion: Sprites-MOT and Dynamics-MOT.

\emph{Sprites-MOT}~\cite{He_TrackingByAnnimation_2019} features 64$\times$64 frames with up to three 11$\times$11 objects, selected from four shapes, moving linearly while entering and leaving the scene. 
This dataset is used to benchmark unsupervised object-centric tracking~\cite{Weis_UnmaskingInductiveBiasesOfUnsupervisedObjectRepresentationsForVideoSequences_2020}.

\emph{Dynamics-MOT} is a self-generated dataset of 30-frame 64$\times$64 videos. Each sequence features three 11$\times$11 objects, selected from a set of eight shapes, moving and bouncing off image boundaries.
To evaluate different motion dynamics, we generate two variants: \emph{Dynamics-MOT Bouncing} features objects moving in linear trajectories, whereas in \emph{Dynamics-MOT Parabolic} the objects follow curved, projectile-like trajectories.
These datasets serve as benchmarks to measure \Method's ability to model more complex object dynamics.

\begin{figure}[tb]
	\centering
	\includegraphics[width=\linewidth]{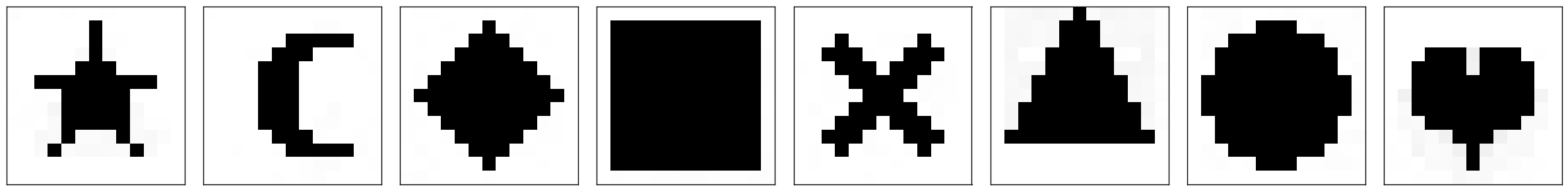}
	\vspace{-0.6cm}
	\caption{
		Learned prototypes on Dynamics-MOT.
		All objects are discovered.
	}
	\label{fig:prot-Dynamics-MOT}
	\vspace{-0.25cm}
\end{figure}

\vspace{0.2cm}

\noindent \textbf{Training: } We train \Method{} in two stages.
First, our framework is trained for object-centric decomposition, enabling the model to learn robust object prototypes and masks.
Figure~\ref{fig:prot-Dynamics-MOT} shows the object prototypes learned from Dynamics-MOT, where \Method{} discovers all eight distinct shapes present in the dataset.
In the second stage, the object prototypes and masks are frozen and the motion module is trained for forecasting future object states---predicting seven future frames given three seed frames.
Overall, \Method{} remains lightweight, with only $~19,200$ learnable parameters.
In \Appendix{app: interpretability}, we analyze and evaluate the interpretability of its internal representations, including object appearance, position and velocity.

\begin{table}[b]
	\centering
	\renewcommand{\arraystretch}{1.}
	\caption{
		Unsupervised object tracking performance on the Sprites-MOT benchmark.
		Our proposed \Method{} model achieves the best tracking performance.
		Best two results are highlighted in boldface and underlined, respectively.
	}
	\vspace*{1ex}
	\begin{tabular}{p{3.0cm} P{1.35cm}P{1.35cm}P{1.1cm}P{1.1cm}P{1.2cm}P{1.02cm}P{1.02cm}}
		\toprule
		 \textbf{Method} & \textbf{MOTA}$\uparrow$ & \textbf{MOTP}$\uparrow$ & \textbf{MD}$\uparrow$ & \textbf{MT}$\uparrow$ & \textbf{Match}$\uparrow$ & \textbf{ID S.}$\downarrow$ & \textbf{FPs}$\downarrow$ \\
		\midrule
		  VideoPCDNet (ours) & \textbf{95.4} & \textbf{94.1} & \underline{94.6} & \underline{92.3} & \textbf{96.1} & 1.0 & \textbf{0.8} \\
		  SCALOR~\cite{Jiang_ScalorGenerativeWorldModelsWithScalableObjectRepresentations_2019} & \underline{94.9} & 80.2 & \textbf{96.4} & \textbf{93.2} & \underline{95.9} & 1.7 & \underline{1.0} \\
		  ViMON\cite{Weis_UnmaskingInductiveBiasesOfUnsupervisedObjectRepresentationsForVideoSequences_2020} & 92.9 & \underline{91.8} & 87.7 & 87.2 & 95.0 & \textbf{0.2} & 2.1 \\
		  OP3~\cite{Veerapaneni_EntityAbstractioninVisualModelBasedReinforcementLearning_2020} & 89.1 & 78.4 & 92.4 & 91.8 & \underline{95.9} & \underline{0.4} & 6.8 \\
		  TBA~\cite{He_TrackingByAnnimation_2019} & 79.7 & 71.2 & 83.4 & 80.0 & 87.8 & 2.6 & 8.1 \\
		  MONet~\cite{Burgess_MonetUnsupervisedSceneDecompositionRepresentation_2019} & 70.2 & 89.6 & 92.4 & 50.4 & 75.3 & 20.3 & 5.1 \\
		\bottomrule
	\end{tabular}
	\label{tab: tracking}
	\vspace{-0.3cm}
\end{table}

\subsection{Unsupervised Object Tracking}

We evaluate \Method{} for unsupervised object tracking on the Sprites-MOT dataset following the evaluation protocol described in~\cite{Weis_UnmaskingInductiveBiasesOfUnsupervisedObjectRepresentationsForVideoSequences_2020} and compare it with multiple object-centric baselines.
The results, listed in Table~\ref{tab: tracking}, show that \Method{} outperforms competing methods, demonstrating a superior tracking accuracy (MOTA) and precision (MOTP).
These findings demonstrate that tracking objects using prototype representations leads to superior accuracy and robustness.

\begin{figure}[b]
	\vspace{0.2cm}
	\centering
	\begin{subfigure}[b]{0.49\linewidth}
		\centering
		\includegraphics[width=\linewidth]{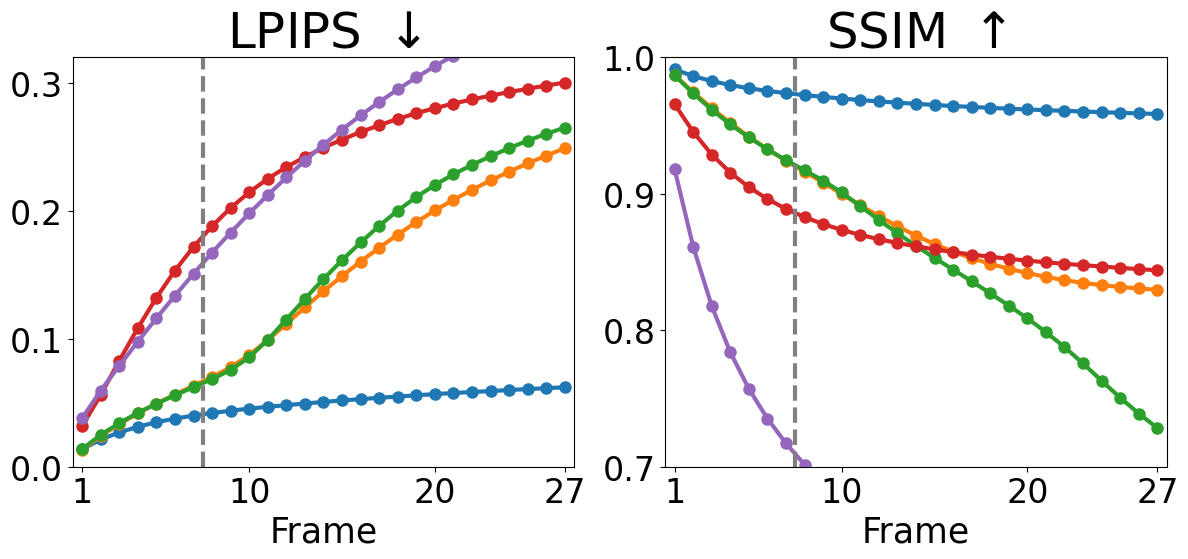}
		\vspace{-0.55cm}
		\caption{Bouncing}
		\label{fig:bouncing}
	\end{subfigure}
	\hfill
	\begin{subfigure}[b]{0.49\linewidth}
		\centering
		\includegraphics[width=\linewidth]{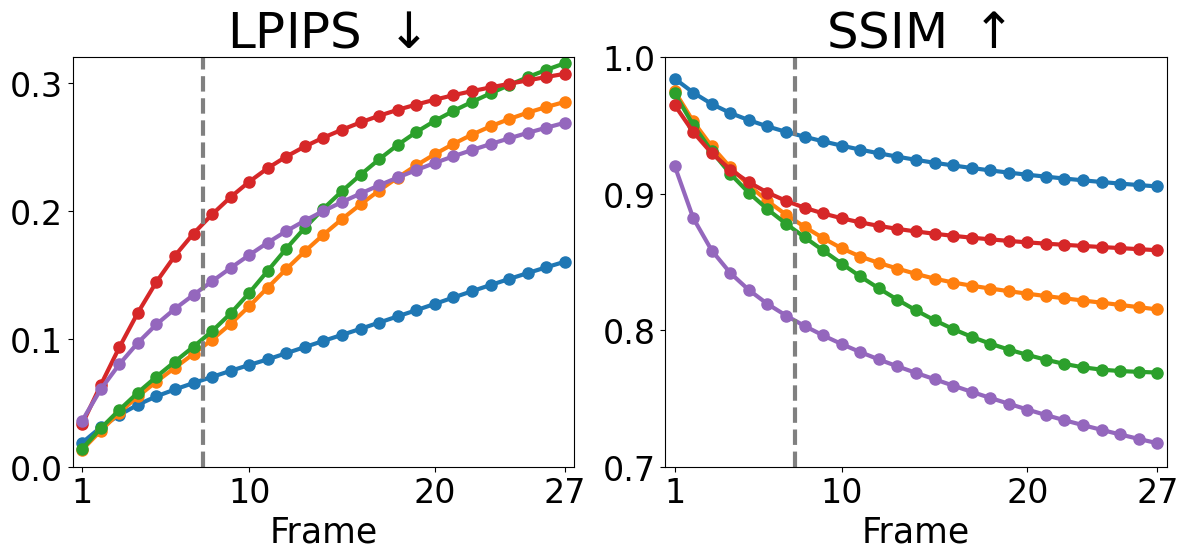}
		\vspace{-0.55cm}
		\caption{Parabolic}
		\label{fig:parabolic}
	\end{subfigure}
	\begin{subfigure}[b]{\linewidth}
		\centering
		\vspace{-0.05cm}
		\includegraphics[width=\linewidth]{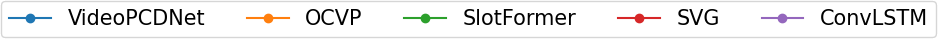}
	\end{subfigure}
	\vspace{-0.5cm}
	\caption{
		Video prediction results on the Dynamics-MOT dataset with (a) Bouncing and (b) Parabolic object motion.
		The vertical bar indicates the prediction horizon used during training.
		\Method{} outperforms all baselines.
	}
	\label{fig:motion_metrics}
	\vspace{-0.3cm}
\end{figure}

\subsection{Ablation Study}
\label{sec:ablation_study}

\begin{table}[t]
	\centering
	\addtolength{\tabcolsep}{-0.025em}
	\renewcommand{\arraystretch}{1.}
	\caption{
		Ablation Study. We evaluate the effect of each component in \Method.
		Best two results are highlighted in boldface and underlined.
	}
	\vspace*{1ex}
	\begin{tabular}{p{0.35cm} p{3.3cm} P{1.39cm}P{1.3cm}P{0.9cm}P{0.95cm}P{1.2cm}P{1.1cm}P{1.cm}}
		\toprule
		& \textbf{Method} & \textbf{MOTA}$\uparrow$ & \textbf{MOTP}$\uparrow$ &  \textbf{MD}$\uparrow$ & \textbf{MT}$\uparrow$ & \textbf{Match}$\uparrow$ & \textbf{ID S.}$\downarrow$ & \textbf{FPs}$\downarrow$ \\
		\midrule
		1 & PCDNet~\cite{Villar_PCDNet_2022} & 77.8 & 73.8 & 90.8 & 65.8 & 80.2 & 15.0 & 2.4 \\
		2 &\hspace{0.15cm} + object align & 93.7 & \underline{95.1} & \underline{95.1} & 88.1 & 94.6 & 2.1 & \underline{1.0} \\
		3 &\hspace{0.15cm} + object state only & 84.8 & \textbf{95.4} & 79.6 & 74.7 & 85.6 & 2.4 & \textbf{0.8} \\   
		4 &\hspace{0.15cm} + two-stage & \underline{94.3} & 94.5 & \textbf{95.9} & \underline{89.1} & \underline{95.3} & \underline{1.9} & \underline{1.0} \\
		5 & \hspace{0.15cm} + external templates & \textbf{95.4} & 94.1 &  94.6 & \textbf{92.3} & \textbf{96.1} & \textbf{1.0} & \textbf{0.8} \\
		\bottomrule
	\end{tabular}
	\label{tab: ablation}
	\vspace{-0.3cm}
\end{table}

We evaluate the tracking performance of different \Method{} variants in order to quantify the effect of each component in our framework.
The results are listed in Table~\ref{tab: ablation}.
\textbf{(1)} the PCDNet baseline, which naively parses each frame individually into its object components, achieves the weakest tracking performance among all variants.
\textbf{(2)} Aligning the object decomposition as described in \Section{sec: state} allows \Method{} to consistently track objects across frames, significantly improving the tracking performance.
\textbf{(3)} Using only the object prototypes present in the state leads to high precision and few false positives. However, it decreases the detection and tracking accuracy as the model cannot correctly represent objects entering the scene.
\textbf{(4)} The two-stage approach described in \Section{sec:dual-stage} addresses this limitation.
\textbf{(5)}  Finally, the full \Method, which additionally incorporates external templates by predicting current object states from past observations, achieves the best tracking performance, demonstrating the effectiveness of each enhancement in boosting overall tracking robustness.

\subsection{Video Prediction}

We further evaluate the capability of \Method{} for modeling object dynamics and predicting future video frames.
We compare our method with two holistic video prediction methods: ConvLSTM~\cite{Shi_ConvLSTMNetworkPrecipitationNowcasting_2015} and  SVG~\cite{Denton_StochasticVideoGenerationWithALearnedPrior_2018}, as well as two object-centric video prediction baselines: SlotFormer~\cite{Wu_SlotFormer_2023} and OCVP~\cite{Villar_OCVP_2023}.

\Figure{fig:motion_metrics} shows the video prediction results on the Dynamics-MOT dataset with Bouncing (\ref{fig:bouncing}) and Parabolic (\ref{fig:parabolic}) object motion, respectively.
Given three seed frames, we predict the subsequent 27 video frames and measure the SSIM and LPIPS metrics.
The results show that object-centric methods, which explicitly model object dynamics (i.e. \Method, SlotFormer and OCVP), outperform their holistic counterparts.
Furthermore, \Method{} consistently achieves the best performance, especially for longer prediction horizons.

\begin{figure}[t]
	\centering
	\begin{subfigure}[b]{0.495\linewidth}
		\centering
		\includegraphics[width=\linewidth]{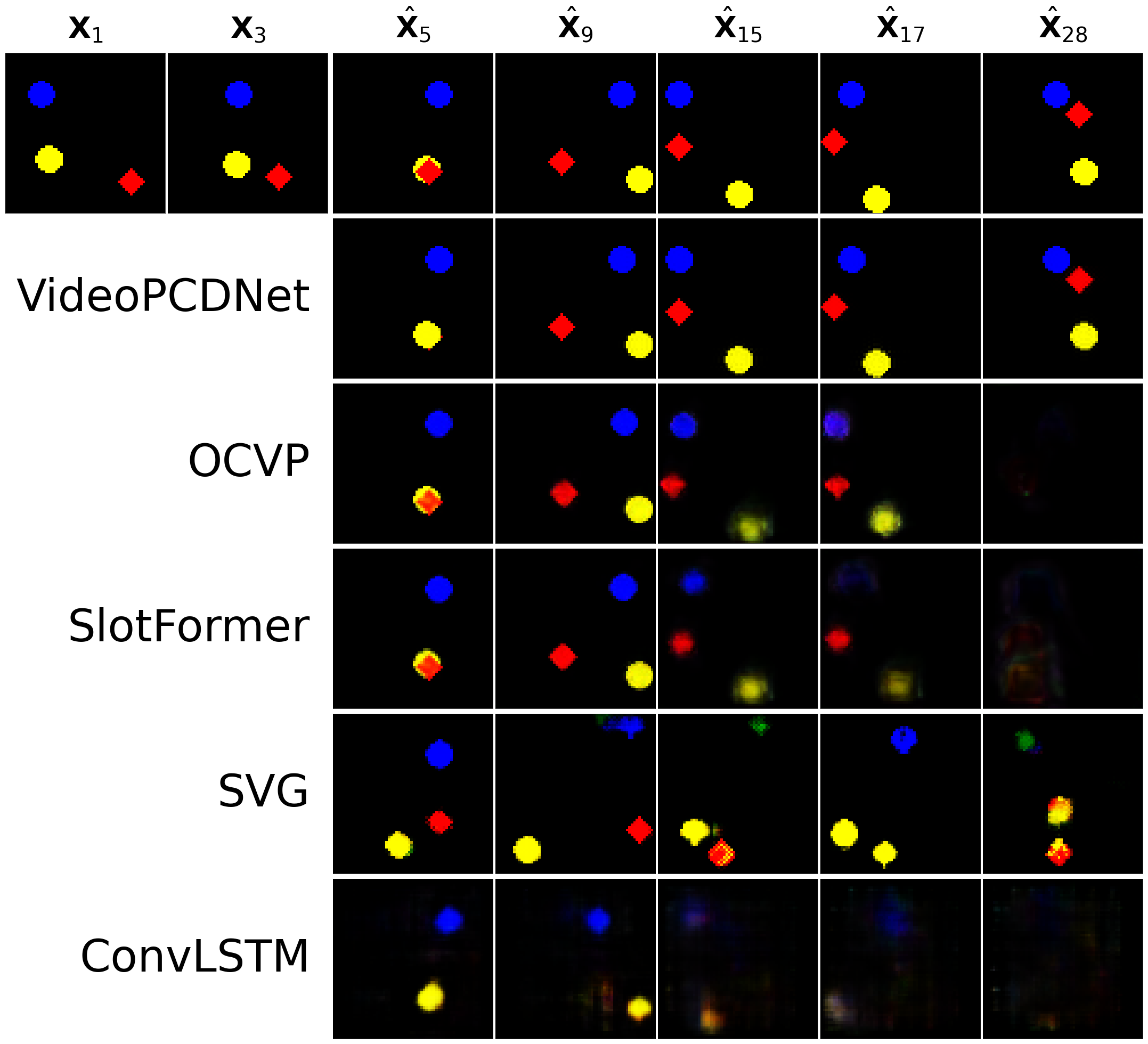}
		\caption{Dynamics-MOT Bouncing}
		\label{fig:pred_bouncing}
	\end{subfigure}
	\hfill
	\begin{subfigure}[b]{0.495\linewidth}
		\centering
		\includegraphics[width=\linewidth]{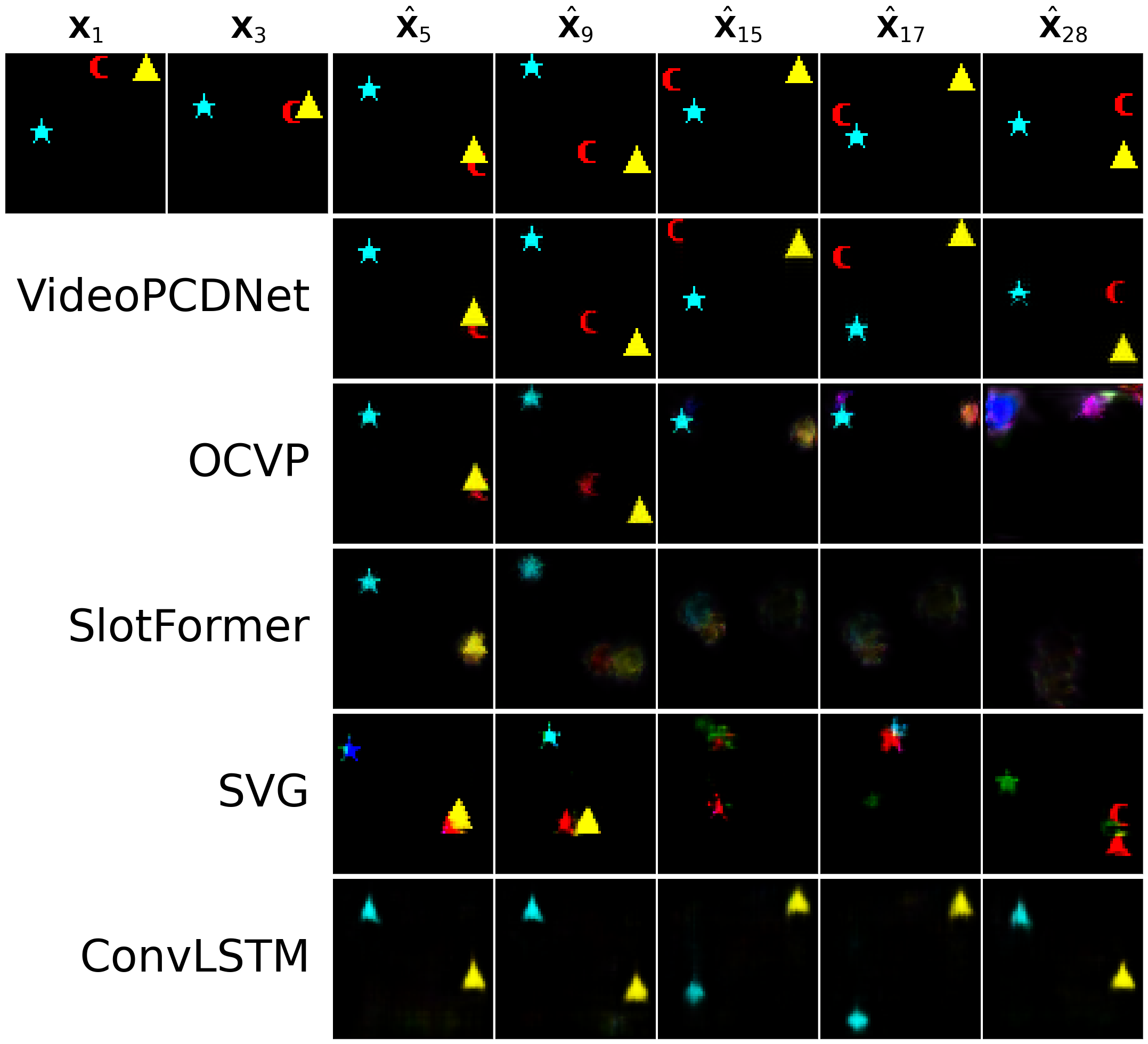}
		\caption{Dynamics-MOT Parabolic}
		\label{fig:pred_parabolic}
	\end{subfigure}
	\vspace{-0.5cm}
	\caption{
		Video prediction rollouts on the Dynamics-MOT dataset with (a) Bouncing and (b) Parabolic object motion.
		\Method{} accurately models the object motion and predicts up to 28 future frames, whereas the baselines suffer from vanishing objects and prediction artifacts.
	}
	\label{fig:predictions}
	\vspace{-0.3cm}
\end{figure}

\Figure{fig:predictions} provides a qualitative comparison illustrating predictions from each method for scenes with bouncing (\ref{fig:pred_bouncing}) and parabolic (\ref{fig:pred_parabolic}) object motion.
In both cases, \Method{} accurately predicts the shape, position, and motion of objects across all time horizons with minimal distortion; whereas the baseline methods progressively degrade in quality, with the non-object-centric SVG and ConvLSTM models exhibiting significant blurriness and visual artifacts in later frames.
These results highlight \Method's capability to handle complex non-linear object motions while preserving detailed visual and structural information throughout extended prediction periods.
Further qualitative video prediction rollouts are shown in \Appendix{app: space inv}.

\section{Conclusion}
We introduced \Method, a novel unsupervised framework for object-centric video parsing and prediction that extends phase-correlation networks to the temporal domain.
\Method{} decomposes video frames into interpretable object components, which are represented as transformed versions from a set of learned object prototypes.
Our model leverages a two-stage approach for recursively parsing a video sequence into their object components, leading to an accurate and robust object tracking even under challenging motion dynamics and occlusions.
Furthermore, \Method{} explicitly models the object motion through phase-correlation techniques in order to efficiently forecast future object states and video frames with less than $20,000$ trainable parameters.
In our evaluations, we demonstrate that \Method{} achieves state-of-the-art performance for unsupervised object tracking on the Sprites-MOT benchmark.
Furthermore, \Method{} accurately forecasts object states over long prediction horizons, outperforming multiple existing video prediction baselines.

\begingroup
\fontsize{9pt}{9pt}\selectfont
	\paragraph{Acknowledgement}
	This work was funded by grant BE 2556/16-2 (Research Unit FOR 2535 Anticipating Human Behavior) of the German Research Foundation (DFG).
\endgroup

\bibliographystyle{splncs04}
\bibliography{referencesAngel}

\clearpage
\appendix

\section{Interpretability}
\label{app: interpretability}

One of the key properties of \Method{} is its ability to parse dynamic scenes into an interpretable representation of object appearances, color, velocity, and position.
In \Section{sec:experimental_setup}, we supported this claim with a qualitative visualization of the learned object prototypes.
In this appendix, we complement these visual insights with further qualitative and quantitative evaluations of interpretability.
We focus on three indicators of interpretability: the quality of predicted object segmentations (\Section{subsec: segm}), the accuracy of predicted object positions (\Section{subsec: pos}), and the visualization of complete object trajectories (\Section{subsec: traj}).

\begin{figure}[b!]
	\centering
	\begin{subfigure}[b]{0.495\linewidth}
		\centering
		\includegraphics[width=\linewidth]{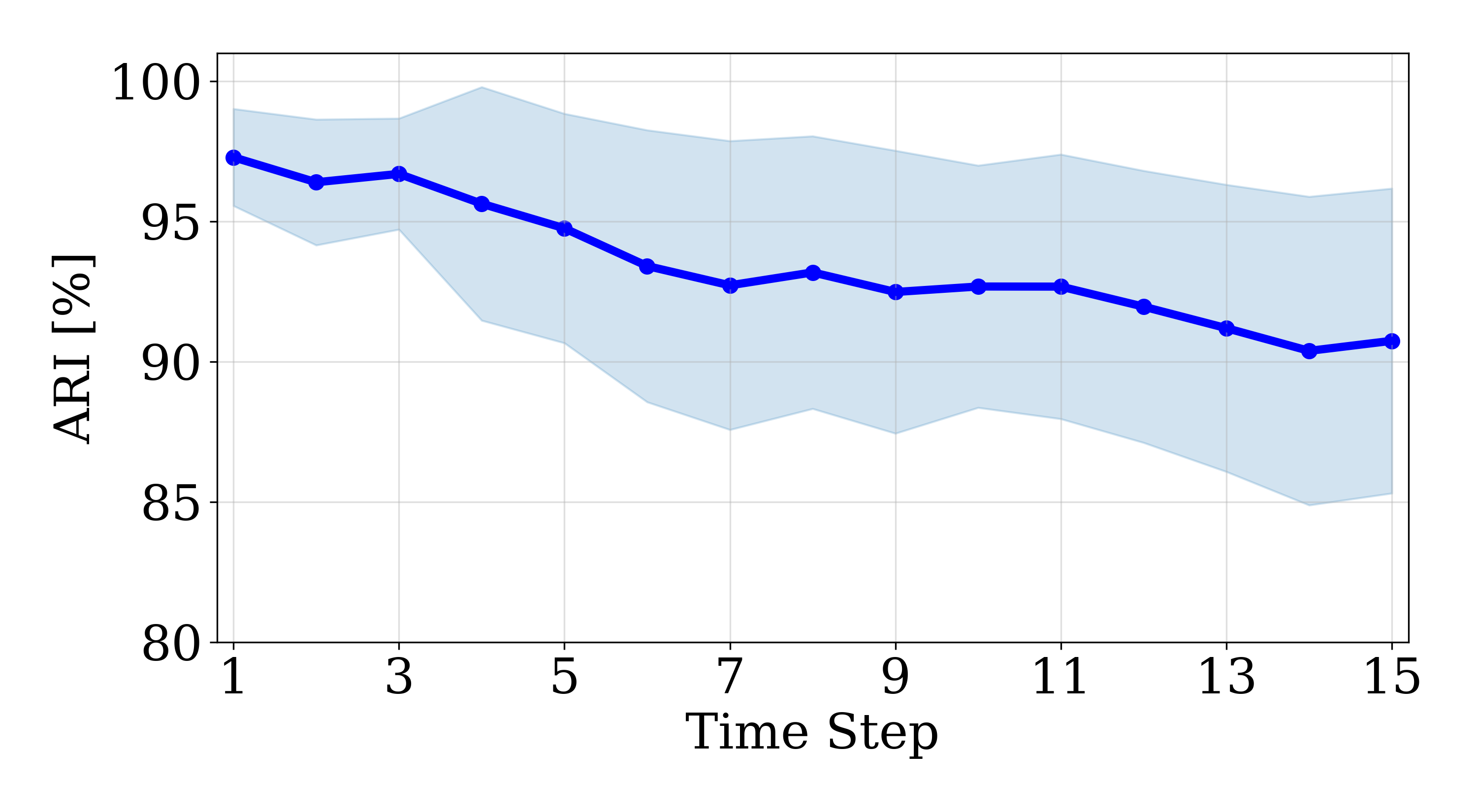}
		\vspace{-0.6cm}
		\caption{Dynamics-MOT Bouncing}
		\label{fig: ari linear}
	\end{subfigure}
	\hfill
	\begin{subfigure}[b]{0.495\linewidth}
		\centering
		\includegraphics[width=\linewidth]{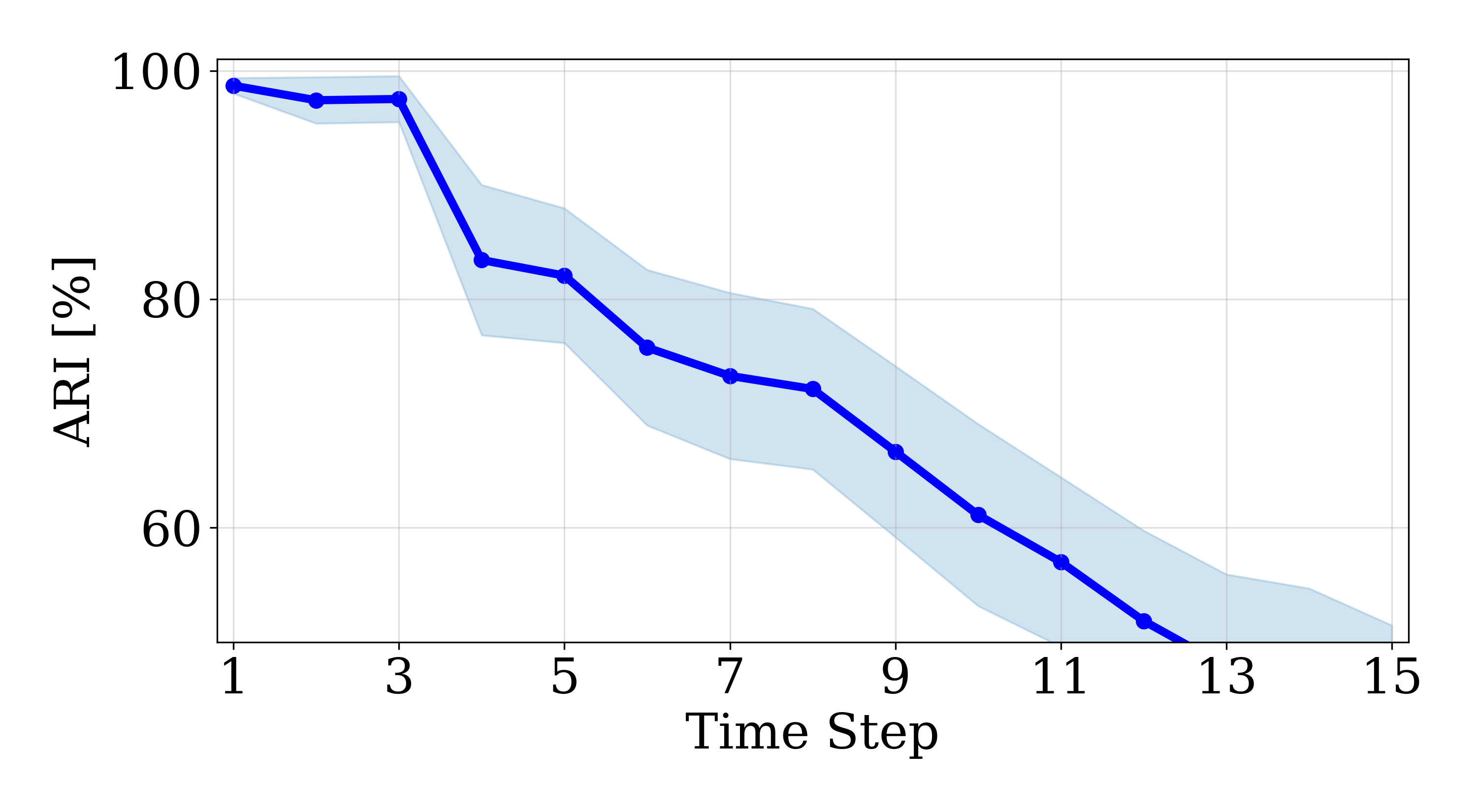}
		\vspace{-0.6cm}
		\caption{Dynamics-MOT Parabolic}
		\label{fig: ari parab}
	\end{subfigure}
	\vspace{-0.6cm}
	\caption{
		Accuracy of predicted object masks over time on the Dynamics-MOT dataset with Bouncing (left) and Parabolic (right) motion.
		Higher ARI indicates better agreement with ground-truth object segmentation.
		\Method{} predicts accurate object masks as shown by the strong segmentation consistency across time-steps.
	}
	\label{fig: ari}
	\vspace{-0.3cm}
\end{figure}

\subsection{Object Segmentations}
\label{subsec: segm}

To quantitatively assess the interpretability of the model’s object-centric representations, we evaluate the accuracy of the predicted object masks produced by \Method{} using the Adjusted Rand Index (ARI)~\cite{Hubert_ComparingPartitionsARI_1985}, computed against the ground-truth segmentation masks.
ARI is a clustering metric that measures the similarity between two set assignments, ignoring label permutations. It ranges from 0 (random assignment) to 1 (perfect segmentation).

Figure~\ref{fig: ari} presents the ARI scores of predicted object masks across prediction time-steps, averaged across 300 test sequences. We show the average ARI score as well as its standard deviation.
We observe that \Method{} generally achieves high ARI scores throughout the prediction horizon, indicating that object shapes are accurately predicted and their identities are consistently preserved.
On the more challenging Dynamics-MOT Parabolic dataset, prediction errors can compound over time, leading to a lower ARI score for longer prediction horizons.
The high ARI scores over time indicate that \Method{} produces accurate and robust object-level decomposition in a predictive setting.

\subsection{Object Position}
\label{subsec: pos}

To further assess the interpretability of the learned representations, we evaluate whether \Method{} can accurately model and track object positions over time.
In the initial decomposition stage, \Method{} estimates object positions by identifying correlation peaks in the phase difference $\Delta \theta$.
During prediction, \Method{} leverages frequency-domain operations and the learned motion module to compute the object velocities $(v^{\text{ref}}_{x}, v^{\text{ref}}_{y} )$, which are then used to forecast future object positions.

\begin{figure}[t]
	\centering
	\begin{subfigure}[b]{0.495\linewidth}
		\centering
		\includegraphics[width=\linewidth]{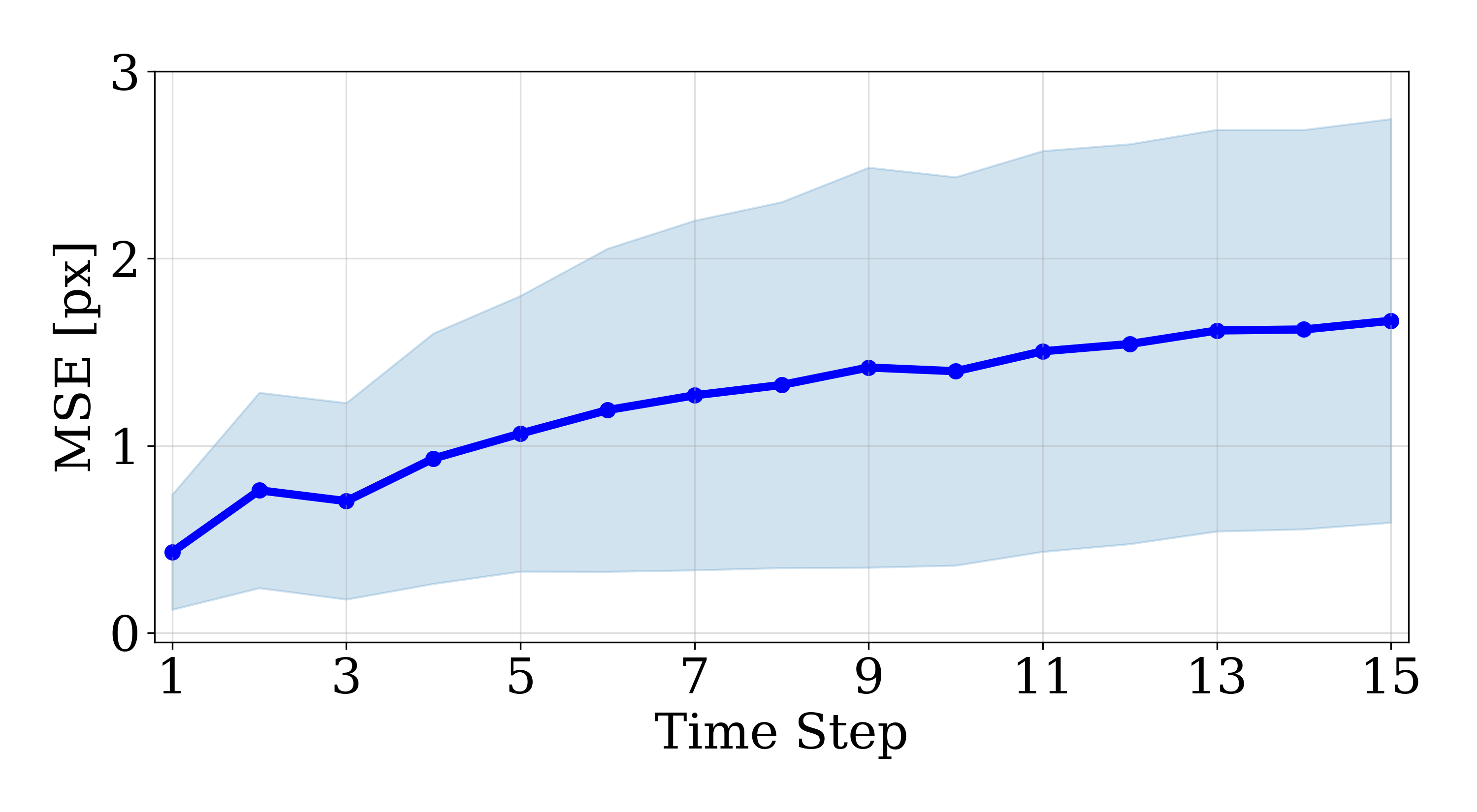}
		\vspace{-0.6cm}
		\caption{Dynamics-MOT Bouncing}
		\label{fig: pos_mse linear}
	\end{subfigure}
	\hfill
	\begin{subfigure}[b]{0.495\linewidth}
		\centering
		\includegraphics[width=\linewidth]{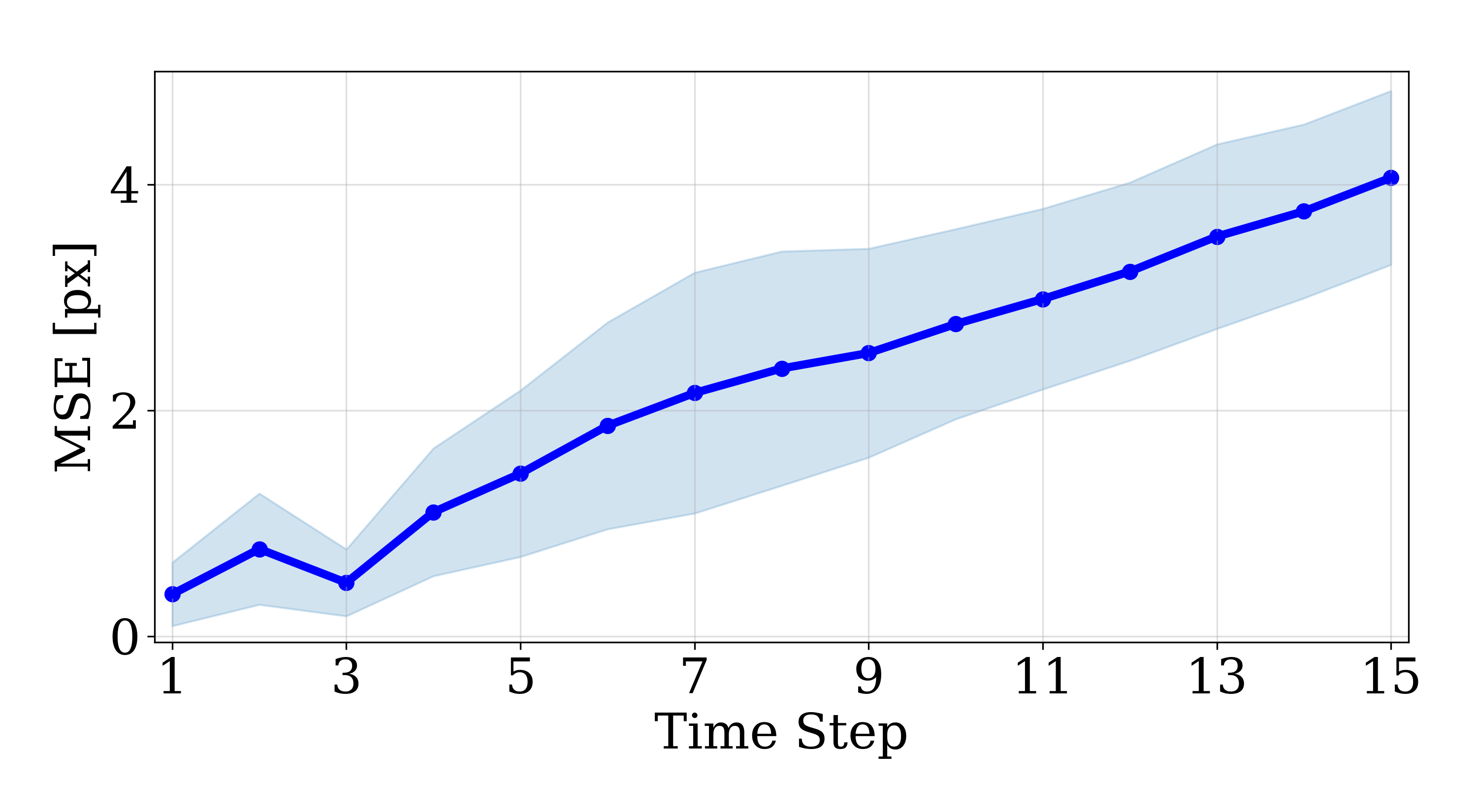}
		\vspace{-0.6cm}
		\caption{Dynamics-MOT Parabolic}
		\label{fig: pos_mse parab}
	\end{subfigure}
	\vspace{-0.6cm}
	\caption{
		Accuracy of predicted object positions over time on the Dynamics-MOT dataset with Bouncing (left) and Parabolic (right) motion.
		\Method{} maintains low position error across future time-steps, demonstrating accurate and consistent object-centric motion prediction in both scenarios.
	}
	\label{fig: pos_mse}
	\vspace{-0.3cm}
\end{figure}

Figure~\ref{fig: pos_mse} reports the mean squared error between predicted and ground-truth object positions, averaged across all objects and 300 sequences.
\Method{} achieves consistently low and stable errors across prediction horizons for both linear and parabolic motion scenarios.
This strong performance---despite the absence of explicit object identity and position supervision---demonstrates that our model learns accurate object-specific position representations.

\subsection{Visualization of Object Trajectories}
\label{subsec: traj}

\begin{figure}[t]
	\centering
	\begin{subfigure}[b]{0.495\linewidth}
		\centering
		\includegraphics[width=\linewidth]{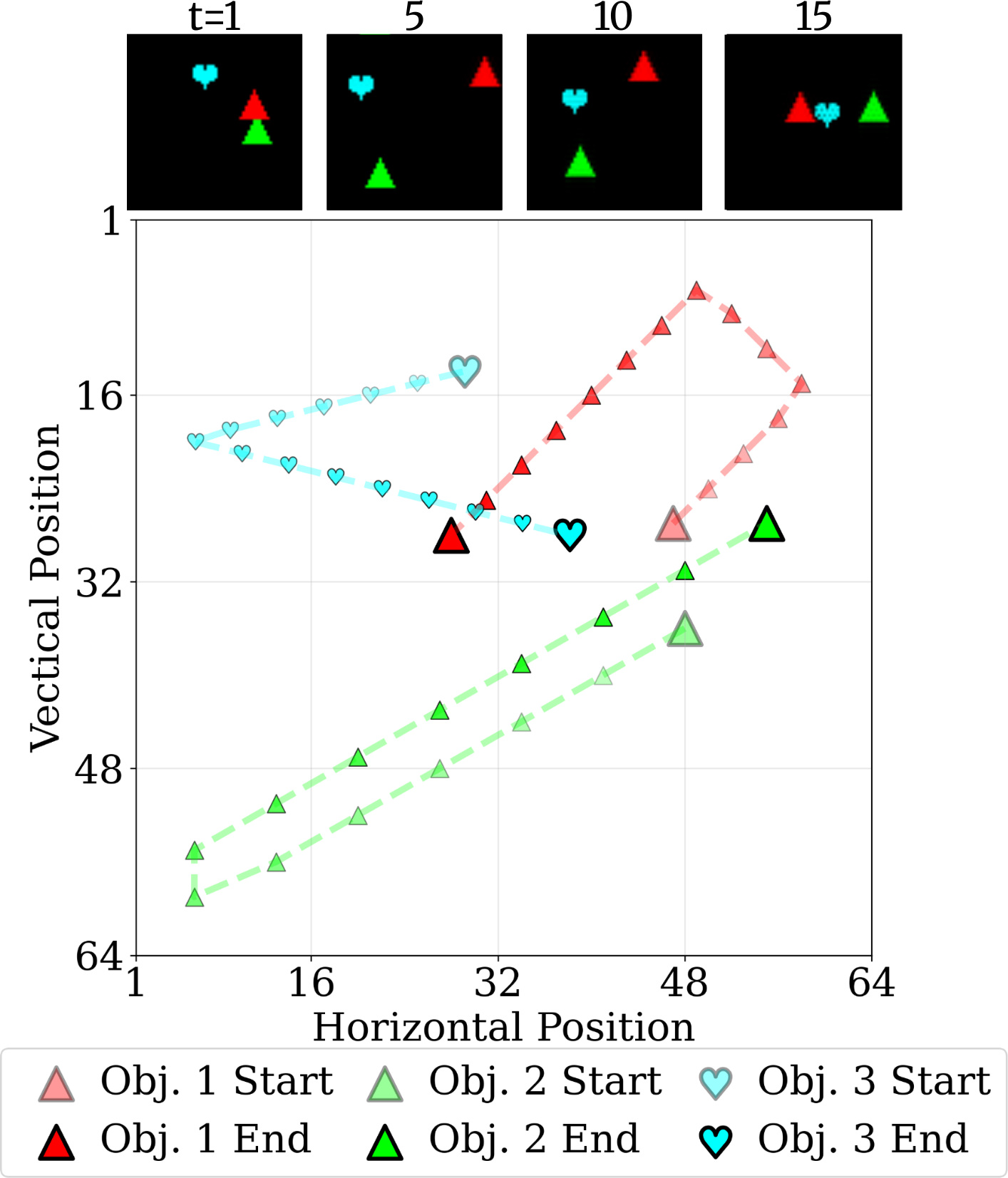}
		\caption{Dynamics-MOT Bouncing}
		\label{fig: traj linear}
	\end{subfigure}
	\hfill
	\begin{subfigure}[b]{0.495\linewidth}
		\centering
		\includegraphics[width=\linewidth]{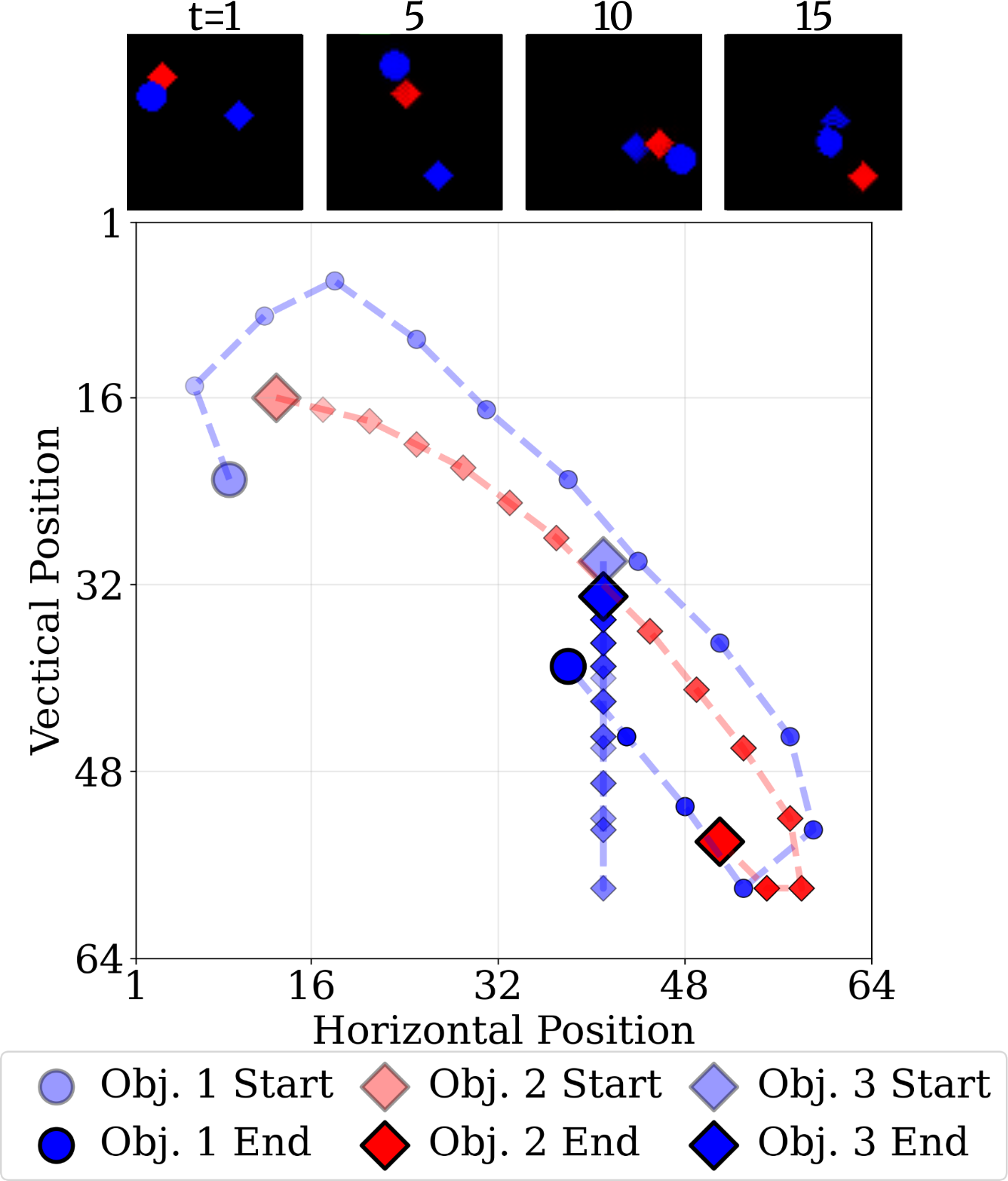}
		\caption{Dynamics-MOT Parabolic}
		\label{fig: traj parab}
	\end{subfigure}
	\vspace{-0.5cm}
	\caption{
		Predicted trajectories on Dynamics-MOT with Bouncing (left) and Parabolic (right) motion.
		\Method{} outputs accurate object trajectories over 15 frames, predicting the corresponding motion patterns and the bouncing behavior when objects collide with image boundaries.
	}
	\label{fig: traj}
	\vspace{-0.2cm}
\end{figure}

To provide a comprehensive view of \Method's interpretability, we visualize the complete object trajectories predicted by our model for a prediction horizon of 15 frames.
\Figure{fig: traj} shows the predicted trajectories for each object in two distinct Dynamics-MOT sequences with linear (\Figure{fig: traj linear}) and parabolic (\Figure{fig: traj parab}) motion.

In both scenarios, \Method{} accurately captures the corresponding motion patterns and correctly predicts the bouncing behavior when objects collide with image boundaries. 
Notably, in the challenging parabolic motion scenario, the predicted trajectories maintain non-linear paths, accurately capturing both the horizontal velocity component and the gravitational acceleration effect.

These results highlight the interpretability of \Method's object-centric approach, where complex scene dynamics are decomposed into meaningful and tractable object trajectories that correspond to intuitive motion patterns.

\section{Evaluation on \SpaceInvaders{} Dataset}
\label{app: space inv}

To demonstrate the applicability of our method to datasets with more visually challenging objects, we evaluate \Method{} on the \SpaceInvaders{} dataset, which presents more complex visual patterns.
We employ two seed frames to predict the subsequent eight.


The \SpaceInvaders{} dataset consists of video sequences mimicking the classic Atari game, featuring multiple aliens and a spaceship.
We render the dataset with six distinct alien sprites, which move from the top of the frame towards the bottom with linear or parabolic velocity, whereas the spaceship moves sideways.
The more complex object appearances and non-linear velocities makes this dataset more challenging that the original Sprites-MOT and Dynamics-MOT benchmarks.

\subsection{Learned Prototypes}
\begin{figure}[t!]
	\centering
	\includegraphics[width=0.99\linewidth]{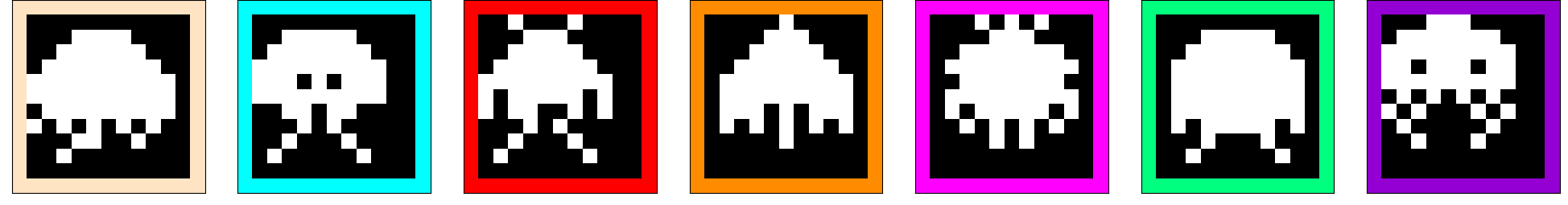}
	\vspace{-0.2cm}
	\caption{Object prototypes learned on the \SpaceInvaders{} dataset.}
	\label{fig: alien protos}
	\vspace{-0.4cm}
\end{figure}

\Figure{fig: alien protos} demonstrates \Method's ability to discover meaningful object prototypes from the \SpaceInvaders{} dataset.
Our model successfully learns seven distinct prototypes, including six different alien shapes and the spaceship.

The diversity of these learned prototypes demonstrates that \Method{} can effectively decompose the dataset into meaningful object-centric representations, extending beyond the simple geometric shapes of synthetic datasets.

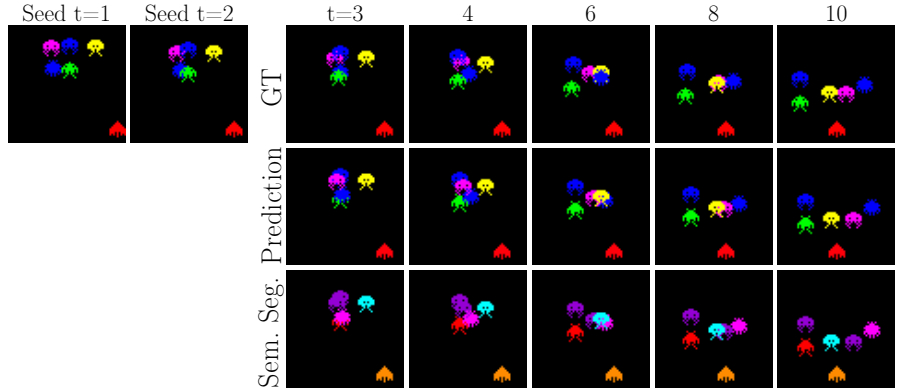
\begin{figure}[b!]
	\centering
	\resizebox{0.98 \linewidth}{!}{
	\begin{tikzpicture} 
		\node(P0)[fill=none] {};
		%
		%
		\node(seed)[anchor=north west, inner sep=0, outer sep=0] at (P0) 
		{\includegraphics[height=3cm]
			{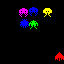}
		};
		\node(seed2)[anchor=west, inner sep=0, outer sep=0] at
		([xshift=0.15cm]seed.east) 
		{\includegraphics[height=3cm]
			{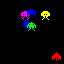}
		};
		\node(target_00)[anchor=west, inner sep=0, outer sep=0] at
		([xshift=1cm]seed2.east) 
		{\includegraphics[height=3cm]
			{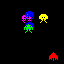}
		};
		\node(0)[anchor=south,draw=none,ultra thick,inner sep=0, outer sep=0] at
		([xshift=-0.4cm, yshift=-2.1cm]target_00.north west)
		{{\rotatebox{90}{\huge GT}}};
		\node(target_01)[anchor=west, inner sep=0, outer sep=0] at
		([xshift=0.15cm]target_00.east) 
		{\includegraphics[height=3cm]
			{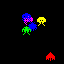}
		};
		\node(target_02)[anchor=west, inner sep=0, outer sep=0] at
		([xshift=0.15cm]target_01.east) 
		{\includegraphics[height=3cm]
			{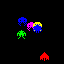}
		};
		\node(target_03)[anchor=west, inner sep=0, outer sep=0] at
		([xshift=0.15cm]target_02.east)
		{\includegraphics[height=3cm]
			{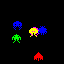}
		};
		\node(target_04)[anchor=west, inner sep=0, outer sep=0] at
		([xshift=0.15cm]target_03.east)
		{\includegraphics[height=3cm]
			{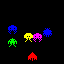}
		};
		%
		%
		%
		%
		%
		%
		%
		%
		\node(pred_00)[anchor=north, inner sep=0, outer sep=0] at
		([xshift=0.00cm, yshift=-0.15cm]target_00.south) 
		{\includegraphics[height=3cm]
			{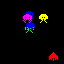}
		};
		\node(0)[anchor=south,draw=none,ultra thick,inner sep=0, outer sep=0] at
		([xshift=-0.4cm, yshift=-3cm]pred_00.north west)
		{{\rotatebox{90}{\huge Prediction}}};
		\node(pred_01)[anchor=north, inner sep=0, outer sep=0] at
		([xshift=0.00cm, yshift=-0.15cm]target_01.south) 
		{\includegraphics[height=3cm]
			{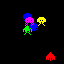}
		};
		\node(pred_02)[anchor=north, inner sep=0, outer sep=0] at
		([xshift=0.00cm, yshift=-0.15cm]target_02.south) 
		{\includegraphics[height=3cm]
			{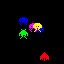}
		};
		\node(pred_03)[anchor=north, inner sep=0, outer sep=0] at
		([xshift=0.00cm, yshift=-0.15cm]target_03.south) 
		{\includegraphics[height=3cm]
			{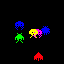}
		};
		\node(pred_04)[anchor=north, inner sep=0, outer sep=0] at
		([xshift=0.00cm, yshift=-0.15cm]target_04.south) 
		{\includegraphics[height=3cm]
			{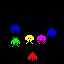}
		};
		%
		%
		%
		%
		%
		%
		%
		%

		%
		\node(pred_00)[anchor=north, inner sep=0, outer sep=0] at
		([xshift=0.00cm, yshift=-0.15cm]pred_00.south) 
		{\includegraphics[height=3cm]
			{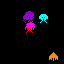}
		};
		\node(0)[anchor=south,draw=none,ultra thick,inner sep=0, outer sep=0] at
		([xshift=-0.4cm, yshift=-3.0cm]pred_00.north west)
		{{\rotatebox{90}{\huge Sem. Seg.}}};
		\node(pred_01)[anchor=north, inner sep=0, outer sep=0] at
		([xshift=0.00cm, yshift=-0.15cm]pred_01.south) 
		{\includegraphics[height=3cm]
			{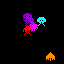}
		};
		\node(pred_02)[anchor=north, inner sep=0, outer sep=0] at
		([xshift=0.00cm, yshift=-0.15cm]pred_02.south) 
		{\includegraphics[height=3cm]
			{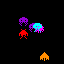}
		};
		\node(pred_03)[anchor=north, inner sep=0, outer sep=0] at
		([xshift=0.00cm, yshift=-0.15cm]pred_03.south) 
		{\includegraphics[height=3cm]
			{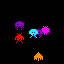}
		};
		\node(pred_04)[anchor=north, inner sep=0, outer sep=0] at
		([xshift=0.00cm, yshift=-0.15cm]pred_04.south) 
		{\includegraphics[height=3cm]
			{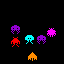}
		};
		%
		%
		%
		%
		\node(0)[anchor=south,draw=none,ultra thick,inner sep=0, outer sep=0] at 	([yshift=0.1cm]seed.north)
		{\LARGE{Seed t=1}};
		\node(0)[anchor=south,draw=none,ultra thick,inner sep=0, outer sep=0] at 	([yshift=0.1cm]seed2.north)
		{\LARGE{Seed t=2}};
		\node(0)[anchor=south,draw=none,ultra thick,inner sep=0, outer sep=0] at 	([yshift=0.1cm]target_00.north)
		{\LARGE{t=3}};
		\node(0)[anchor=south,draw=none,ultra thick,inner sep=0, outer sep=0] at 	([yshift=0.1cm]target_01.north)
		{\LARGE{4}};
		\node(0)[anchor=south,draw=none,ultra thick,inner sep=0, outer sep=0] at 	([yshift=0.1cm]target_02.north)
		{\LARGE{6}};
		\node(0)[anchor=south,draw=none,ultra thick,inner sep=0, outer sep=0] at 	([yshift=0.1cm]target_03.north)
		{\LARGE{8}};
		\node(0)[anchor=south,draw=none,ultra thick,inner sep=0, outer sep=0] at 	([yshift=0.1cm]target_04.north)
		{\LARGE{10}};
	\end{tikzpicture}
}
\vspace{-0.15cm}
\caption{
	Qualitative results on \SpaceInvaders{} dataset showing ground truth with two seed frames (top), \Method{} predictions (middle), and semantic segmentation (bottom, colors from \Figure{fig: alien protos}).
}
\label{fig: aliens_00}
\end{figure}

\subsection{Video Parsing and Prediction}
\Figuress{fig: aliens_00}{fig: aliens_07} provide qualitative results demonstrating \Method's video parsing and prediction capabilities across different sequences from the \SpaceInvaders{} dataset.
Each figure shows \Method's predicted frames and the corresponding predicted semantic segmentation masks.
In all sequences, our model successfully maintains object identity and appearance while accurately predicting their trajectories.
Furthermore, the semantic segmentation masks demonstrate that \Method{} preserves sharp object boundaries and distinct object identities throughout the prediction horizon, even as objects move across the scene, suffer from overlap, or interact with boundaries.

\begin{figure}[b!]
	\centering
	\resizebox{0.98 \linewidth}{!}{
		\begin{tikzpicture} 
			\node(P0)[fill=none] {};
			%
			%
			\node(seed)[anchor=north west, inner sep=0, outer sep=0] at (P0) 
			{\includegraphics[height=3cm]
				{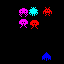}
			};
			\node(seed2)[anchor=west, inner sep=0, outer sep=0] at
			([xshift=0.15cm]seed.east) 
			{\includegraphics[height=3cm]
				{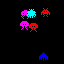}
			};
			\node(target_00)[anchor=west, inner sep=0, outer sep=0] at
			([xshift=1cm]seed2.east) 
			{\includegraphics[height=3cm]
				{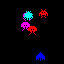}
			};
			\node(0)[anchor=south,draw=none,ultra thick,inner sep=0, outer sep=0] at
			([xshift=-0.4cm, yshift=-2.1cm]target_00.north west)
			{{\rotatebox{90}{\huge GT}}};
			\node(target_01)[anchor=west, inner sep=0, outer sep=0] at
			([xshift=0.15cm]target_00.east) 
			{\includegraphics[height=3cm]
				{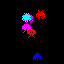}
			};
			\node(target_02)[anchor=west, inner sep=0, outer sep=0] at
			([xshift=0.15cm]target_01.east) 
			{\includegraphics[height=3cm]
				{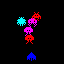}
			};
			\node(target_03)[anchor=west, inner sep=0, outer sep=0] at
			([xshift=0.15cm]target_02.east)
			{\includegraphics[height=3cm]
				{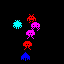}
			};
			\node(target_04)[anchor=west, inner sep=0, outer sep=0] at
			([xshift=0.15cm]target_03.east)
			{\includegraphics[height=3cm]
				{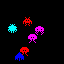}
			};
			%
			%
			%
			%
			%
			%
			%
			%
			%
			\node(pred_00)[anchor=north, inner sep=0, outer sep=0] at
			([xshift=0.00cm, yshift=-0.15cm]target_00.south) 
			{\includegraphics[height=3cm]
				{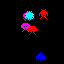}
			};
			\node(0)[anchor=south,draw=none,ultra thick,inner sep=0, outer sep=0] at
			([xshift=-0.4cm, yshift=-3cm]pred_00.north west)
			{{\rotatebox{90}{\huge Prediction}}};
			\node(pred_01)[anchor=north, inner sep=0, outer sep=0] at
			([xshift=0.00cm, yshift=-0.15cm]target_01.south) 
			{\includegraphics[height=3cm]
				{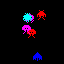}
			};
			\node(pred_02)[anchor=north, inner sep=0, outer sep=0] at
			([xshift=0.00cm, yshift=-0.15cm]target_02.south) 
			{\includegraphics[height=3cm]
				{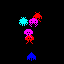}
			};
			\node(pred_03)[anchor=north, inner sep=0, outer sep=0] at
			([xshift=0.00cm, yshift=-0.15cm]target_03.south) 
			{\includegraphics[height=3cm]
				{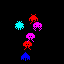}
			};
			\node(pred_04)[anchor=north, inner sep=0, outer sep=0] at
			([xshift=0.00cm, yshift=-0.15cm]target_04.south) 
			{\includegraphics[height=3cm]
				{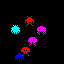}
			};
			%
			%
			%
			%
			%
			%
			%
			%
			\node(pred_00)[anchor=north, inner sep=0, outer sep=0] at
			([xshift=0.00cm, yshift=-0.15cm]pred_00.south) 
			{\includegraphics[height=3cm]
				{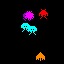}
			};
			\node(0)[anchor=south,draw=none,ultra thick,inner sep=0, outer sep=0] at
			([xshift=-0.4cm, yshift=-3.0cm]pred_00.north west)
			{{\rotatebox{90}{\huge Sem. Seg.}}};
			\node(pred_01)[anchor=north, inner sep=0, outer sep=0] at
			([xshift=0.00cm, yshift=-0.15cm]pred_01.south) 
			{\includegraphics[height=3cm]
				{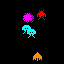}
			};
			\node(pred_02)[anchor=north, inner sep=0, outer sep=0] at
			([xshift=0.00cm, yshift=-0.15cm]pred_02.south) 
			{\includegraphics[height=3cm]
				{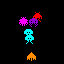}
			};
			\node(pred_03)[anchor=north, inner sep=0, outer sep=0] at
			([xshift=0.00cm, yshift=-0.15cm]pred_03.south) 
			{\includegraphics[height=3cm]
				{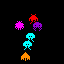}
			};
			\node(pred_04)[anchor=north, inner sep=0, outer sep=0] at
			([xshift=0.00cm, yshift=-0.15cm]pred_04.south) 
			{\includegraphics[height=3cm]
				{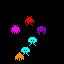}
			};
			%
			%
			%
			%
			\node(0)[anchor=south,draw=none,ultra thick,inner sep=0, outer sep=0] at 	([yshift=0.1cm]seed.north)
			{\LARGE{Seed t=1}};
			\node(0)[anchor=south,draw=none,ultra thick,inner sep=0, outer sep=0] at 	([yshift=0.1cm]seed2.north)
			{\LARGE{Seed t=2}};
			\node(0)[anchor=south,draw=none,ultra thick,inner sep=0, outer sep=0] at 	([yshift=0.1cm]target_00.north)
			{\LARGE{t=3}};
			\node(0)[anchor=south,draw=none,ultra thick,inner sep=0, outer sep=0] at 	([yshift=0.1cm]target_01.north)
			{\LARGE{4}};
			\node(0)[anchor=south,draw=none,ultra thick,inner sep=0, outer sep=0] at 	([yshift=0.1cm]target_02.north)
			{\LARGE{6}};
			\node(0)[anchor=south,draw=none,ultra thick,inner sep=0, outer sep=0] at 	([yshift=0.1cm]target_03.north)
			{\LARGE{8}};
			\node(0)[anchor=south,draw=none,ultra thick,inner sep=0, outer sep=0] at 	([yshift=0.1cm]target_04.north)
			{\LARGE{10}};
		\end{tikzpicture}
	}
	\vspace{-0.15cm}
	\caption{
		Qualitative results on \SpaceInvaders{} dataset showing ground truth with two seed frames (top), \Method{} predictions (middle), and semantic segmentation (bottom, colors from \Figure{fig: alien protos}).
	}
	\label{fig: aliens_04}
\end{figure}

\begin{figure}[b!]
	\centering
	\resizebox{0.98 \linewidth}{!}{
		\begin{tikzpicture} 
			\node(P0)[fill=none] {};
			%
			%
			\node(seed)[anchor=north west, inner sep=0, outer sep=0] at (P0) 
			{\includegraphics[height=3cm]
				{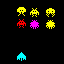}
			};
			\node(seed2)[anchor=west, inner sep=0, outer sep=0] at
			([xshift=0.15cm]seed.east) 
			{\includegraphics[height=3cm]
				{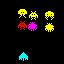}
			};
			\node(target_00)[anchor=west, inner sep=0, outer sep=0] at
			([xshift=1cm]seed2.east) 
			{\includegraphics[height=3cm]
				{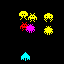}
			};
			\node(0)[anchor=south,draw=none,ultra thick,inner sep=0, outer sep=0] at
			([xshift=-0.4cm, yshift=-2.1cm]target_00.north west)
			{{\rotatebox{90}{\huge GT}}};
			\node(target_01)[anchor=west, inner sep=0, outer sep=0] at
			([xshift=0.15cm]target_00.east) 
			{\includegraphics[height=3cm]
				{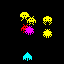}
			};
			\node(target_02)[anchor=west, inner sep=0, outer sep=0] at
			([xshift=0.15cm]target_01.east) 
			{\includegraphics[height=3cm]
				{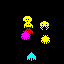}
			};
			\node(target_03)[anchor=west, inner sep=0, outer sep=0] at
			([xshift=0.15cm]target_02.east)
			{\includegraphics[height=3cm]
				{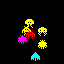}
			};
			\node(target_04)[anchor=west, inner sep=0, outer sep=0] at
			([xshift=0.15cm]target_03.east)
			{\includegraphics[height=3cm]
				{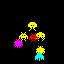}
			};
			%
			%
			%
			%
			%
			%
			%
			%
			%
			\node(pred_00)[anchor=north, inner sep=0, outer sep=0] at
			([xshift=0.00cm, yshift=-0.15cm]target_00.south) 
			{\includegraphics[height=3cm]
				{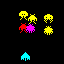}
			};
			\node(0)[anchor=south,draw=none,ultra thick,inner sep=0, outer sep=0] at
			([xshift=-0.4cm, yshift=-3cm]pred_00.north west)
			{{\rotatebox{90}{\huge Prediction}}};
			\node(pred_01)[anchor=north, inner sep=0, outer sep=0] at
			([xshift=0.00cm, yshift=-0.15cm]target_01.south) 
			{\includegraphics[height=3cm]
				{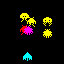}
			};
			\node(pred_02)[anchor=north, inner sep=0, outer sep=0] at
			([xshift=0.00cm, yshift=-0.15cm]target_02.south) 
			{\includegraphics[height=3cm]
				{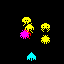}
			};
			\node(pred_03)[anchor=north, inner sep=0, outer sep=0] at
			([xshift=0.00cm, yshift=-0.15cm]target_03.south) 
			{\includegraphics[height=3cm]
				{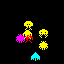}
			};
			\node(pred_04)[anchor=north, inner sep=0, outer sep=0] at
			([xshift=0.00cm, yshift=-0.15cm]target_04.south) 
			{\includegraphics[height=3cm]
				{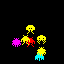}
			};
			%
			%
			%
			%
			%
			%
			%
			%
			%
			\node(pred_00)[anchor=north, inner sep=0, outer sep=0] at
			([xshift=0.00cm, yshift=-0.15cm]pred_00.south) 
			{\includegraphics[height=3cm]
				{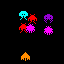}
			};
			\node(0)[anchor=south,draw=none,ultra thick,inner sep=0, outer sep=0] at
			([xshift=-0.4cm, yshift=-3.0cm]pred_00.north west)
			{{\rotatebox{90}{\huge Sem. Seg.}}};
			\node(pred_01)[anchor=north, inner sep=0, outer sep=0] at
			([xshift=0.00cm, yshift=-0.15cm]pred_01.south) 
			{\includegraphics[height=3cm]
				{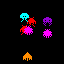}
			};
			\node(pred_02)[anchor=north, inner sep=0, outer sep=0] at
			([xshift=0.00cm, yshift=-0.15cm]pred_02.south) 
			{\includegraphics[height=3cm]
				{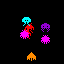}
			};
			\node(pred_03)[anchor=north, inner sep=0, outer sep=0] at
			([xshift=0.00cm, yshift=-0.15cm]pred_03.south) 
			{\includegraphics[height=3cm]
				{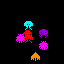}
			};
			\node(pred_04)[anchor=north, inner sep=0, outer sep=0] at
			([xshift=0.00cm, yshift=-0.15cm]pred_04.south) 
			{\includegraphics[height=3cm]
				{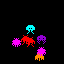}
			};
			%
			%
			%
			%
			\node(0)[anchor=south,draw=none,ultra thick,inner sep=0, outer sep=0] at 	([yshift=0.1cm]seed.north)
			{\LARGE{Seed t=1}};
			\node(0)[anchor=south,draw=none,ultra thick,inner sep=0, outer sep=0] at 	([yshift=0.1cm]seed2.north)
			{\LARGE{Seed t=2}};
			\node(0)[anchor=south,draw=none,ultra thick,inner sep=0, outer sep=0] at 	([yshift=0.1cm]target_00.north)
			{\LARGE{t=3}};
			\node(0)[anchor=south,draw=none,ultra thick,inner sep=0, outer sep=0] at 	([yshift=0.1cm]target_01.north)
			{\LARGE{4}};
			\node(0)[anchor=south,draw=none,ultra thick,inner sep=0, outer sep=0] at 	([yshift=0.1cm]target_02.north)
			{\LARGE{6}};
			\node(0)[anchor=south,draw=none,ultra thick,inner sep=0, outer sep=0] at 	([yshift=0.1cm]target_03.north)
			{\LARGE{8}};
			\node(0)[anchor=south,draw=none,ultra thick,inner sep=0, outer sep=0] at 	([yshift=0.1cm]target_04.north)
			{\LARGE{10}};
		\end{tikzpicture}
	}
	\vspace{-0.15cm}
	\caption{
		Qualitative results on \SpaceInvaders{} dataset showing ground truth with two seed frames (top), \Method{} predictions (middle), and semantic segmentation (bottom, colors from \Figure{fig: alien protos}).
	}
	\label{fig: aliens_07}
\end{figure}

\end{document}